\documentclass[11pt,a4paper]{article}
\usepackage[margin=2.5cm]{geometry}
\usepackage[utf8]{inputenc}
\usepackage[T1]{fontenc}
\usepackage{textcomp}
\usepackage{mathptmx}
\usepackage{amsmath,amssymb}
\usepackage{newunicodechar}
\newunicodechar{−}{\ifmmode-\else\textminus\fi}
\newunicodechar{γ}{\ensuremath{\gamma}}
\newunicodechar{α}{\ensuremath{\alpha}}
\newunicodechar{π}{\ensuremath{\pi}}
\newunicodechar{δ}{\ensuremath{\delta}}
\newunicodechar{Δ}{\ensuremath{\Delta}}
\newunicodechar{τ}{\ensuremath{\tau}}
\newunicodechar{²}{\ensuremath{^{2}}}
\newunicodechar{³}{\ensuremath{^{3}}}
\newunicodechar{⁴}{\ensuremath{^{4}}}
\newunicodechar{≥}{\ensuremath{\geq}}
\usepackage{longtable,booktabs,array,calc}
\usepackage{graphicx}
\usepackage{needspace}
\usepackage{xcolor}
\usepackage[hidelinks]{hyperref}
\usepackage{xurl}
\usepackage{microtype}
\usepackage{setspace}
\makeatletter
\def\maxwidth{\ifdim\Gin@nat@width>\linewidth\linewidth\else\Gin@nat@width\fi}
\def\maxheight{\ifdim\Gin@nat@height>0.45\textheight 0.45\textheight\else\Gin@nat@height\fi}
\makeatother
\setkeys{Gin}{width=\maxwidth,height=\maxheight,keepaspectratio}

\title{Distribution-Free Uncertainty Quantification for Weather Forecasting:\\ Conformal Prediction Under Spatiotemporal Dependence}
\author{Mariya\\ Stanley College of Engineering and Technology for Women, Hyderabad, India}
\date{}
\begin{document}
\maketitle
\begin{abstract}
Conformal prediction gives finite-sample, distribution-free marginal coverage and has begun to be applied to gridded weather forecasts. Marginal coverage concerns the long-run fraction of covered points at a location, not how coverage behaves jointly across space and time on a given day. I benchmark eight interval-construction procedures (Gaussian, pooled and per-cell split conformal, per-season, adaptive conformal inference and rolling windows) for gridded 2 m temperature forecasts at eight lead times from 6 hours to 10 days, using ERA5 at 5.625° resolution, a deliberately modest convolutional forecaster and ECMWF's operational HRES, with calibration in 2017--2018 and testing in 2019--2021. Alongside coverage, width and interval score, I report dependence diagnostics: a spatial overdispersion index, quantiles of daily spatial coverage and miss persistence relative to a shuffled-days null. Per-cell methods reach near-nominal marginal coverage, yet the day-to-day variance of the fraction of cells covered is roughly 3 to 10 times what independent cells would give, and misses cluster in time; both effects grow with lead time for the CNN, also appear for HRES, and are shared by all methods. An oracle that divides each day's normalised errors by that day's mean removes 81 to 92 percent of the excess dispersion for the CNN and 93 to 98 percent for HRES, so days of large errors across much of the grid account for most of it. A day-scale predicted from seasonal terms and already-verified errors lowers the overdispersion index without widening intervals but improves interval score by at most 0.6 percent; on HRES from 12 hours to 3 days it has the lowest overdispersion index while ACI or season-conditional calibration has the best interval score, so no method is best on both. The study covers one variable, one coarse grid, a reanalysis target and two forecasters.
\end{abstract}

\hypertarget{introduction}{%
\section{Introduction}\label{introduction}}

Reliable uncertainty estimates are central to the use of weather forecasts. Operational centres obtain them from ensembles of numerical simulations, whereas many recent data-driven forecasters, of the kind compared in the WeatherBench benchmarks {[}14, 15{]}, are trained and evaluated as deterministic predictors. Distribution-free post-hoc methods such as conformal prediction {[}1, 2, 3, 4{]} are an attractive way to attach intervals to any point forecast: under exchangeability, split conformal prediction guarantees finite-sample marginal coverage regardless of the forecaster or the error distribution.

For gridded forecasts, however, this guarantee is easy to over-read. It concerns the long-run frequency with which the interval at a given location and lead time contains the truth. It does not say how the fraction of covered grid cells behaves on an individual day. If forecast errors are spatially correlated, as large-scale atmospheric anomalies make them, a nominal 90\% interval can cover the nominal fraction of cells on average while covering noticeably fewer on some days and more on others. A user of a single forecast field experiences the daily outcome, not the average. The same issue arises in time: if a miss at one initialisation makes a miss at the next more likely, intervals fail in runs rather than independently. The problem is analogous to field significance in climate science, where spatial dependence changes the effective number of independent tests {[}23, 24{]}.

Temporal structure creates a second complication for online procedures. Adaptive conformal inference {[}5{]} updates its miscoverage level using the outcome of each past prediction and, in its original form, assumes that outcome is revealed before the next prediction is made. A forecast at lead τ cannot be verified until τ hours after it is issued, so at long leads the feedback available to an online method is stale by days.

\hypertarget{contributions.}{%
\paragraph{Contributions.}\label{contributions.}}

\begin{itemize}
\item
  I benchmark eight interval procedures, including adaptive and rolling-window methods run with realistic delayed feedback, on global gridded 2\,m temperature forecasts at eight lead times (6\,h to 10\,d) under a strict temporal split.
\item
  I adapt and validate a set of dependence-aware diagnostics: an area-weighted spatial overdispersion index, quantiles of the daily fraction of cells covered, miss persistence relative to a shuffled-days null, and coverage by season and test year, with block-bootstrap intervals and paired comparisons between methods.
\item
  I report what these diagnostics show: marginal coverage is close to nominal for per-cell methods and shows no clear drift over three test years, but spatial overdispersion and temporal clustering of misses are large, grow with lead time, and are shared by all methods.
\item
  I decompose the excess dispersion: an oracle day-scale diagnostic, with a null-floor control and a split-half variant, shows that most of it is a broad day-level effect (Section~4.8).
\item
  I add a second, skilful forecaster (ECMWF HRES) and show that the dependence effects are not an artifact of a weak forecaster, while noting that HRES is not a fixed system (Sections~4.10 and 4.11).
\item
  I test whether the day-scale can be predicted when the forecast is issued. The predictable part is mainly seasonal and short-lead persistence, and interval-score gains are at most 0.6 percent, and on HRES no method is best on both day-level reliability and interval score from 12 hours to 3 days (Sections~4.9 and 4.11).
\item
  I state the limits of the evidence and outline a dependence-aware method motivated by the failure mode I observe (Section~7); I test only a minimal, global form of its first element (Sections~4.8 to 4.11).
\end{itemize}

I propose no new conformal method in this paper. My aim is to establish, with honest uncertainty estimates, which properties of standard procedures hold in this setting and which do not.

\hypertarget{related-work}{%
\section{Related work}\label{related-work}}

\hypertarget{conformal-prediction-and-its-extensions.}{%
\paragraph{Conformal prediction and its extensions.}\label{conformal-prediction-and-its-extensions.}}

Split (inductive) conformal prediction {[}2, 3{]} and its theory {[}1{]} yield intervals with finite-sample marginal coverage under exchangeability; {[}4{]} gives a recent introduction. Weather and climate data violate exchangeability through seasonality, serial correlation and trends. Work on non-exchangeable data includes coverage bounds that degrade with a measure of distribution shift {[}7{]}, split-conformal validity for many non-exchangeable processes at the price of a small coverage penalty {[}10{]}, adaptive conformal inference (ACI), which adjusts the working miscoverage level online {[}5{]}, aggregation of ACI step sizes {[}6{]}, ensemble-based intervals for dynamic time series {[}8{]}, and intervals designed for temporally dependent data {[}9{]}. My online baselines are ACI-style per-cell updates and rolling-window recalibration; I evaluate them with delayed feedback, which matches how forecasts are verified.

\hypertarget{spatial-and-multivariate-outputs.}{%
\paragraph{Spatial and multivariate outputs.}\label{spatial-and-multivariate-outputs.}}

Model-free spatial prediction with conformal methods is studied in {[}11{]}, and multi-target regression with dependence between outputs in, for example, {[}12{]}. Spatio-temporal conformal inference based on local exchangeability is developed in {[}26{]}, and {[}10{]} reports experiments with split conformal prediction on time series and spatiotemporal data. My setting is different in emphasis: I keep a standard per-cell construction and ask how coverage varies across cells and days when errors are spatially and temporally dependent.

\hypertarget{conformal-prediction-for-weather-and-climate.}{%
\paragraph{Conformal prediction for weather and climate.}\label{conformal-prediction-for-weather-and-climate.}}

Conformal prediction has recently been applied to weather and climate. Gopakumar et al. {[}17{]} apply cell-wise split-conformal calibration to spatio-temporal surrogate models, including a weather forecasting application, with a fixed adjustment that assumes exchangeability. Asch et al. {[}25{]} apply online adaptive conformal inference per grid point, lead time and quantile to ensemble forecasts from GenCast, NeuralGCM and AIFS-ENS, and note that separate per-location adjustments break spatial covariance, leaving dependence-aware calibration to future work. Brusaferri and Ballarino {[}18{]} conformalise a latent diffusion model for kilometre-scale downscaling of ERA5 over Italy, and Harris and Sriver {[}19{]} build conformal ensembles of climate projections. These works report marginal or grid-point-level coverage. Classical ensemble post-processing such as EMOS {[}27{]} corrects marginal calibration, while the Schaake shuffle {[}28{]} and ensemble copula coupling {[}29{]} restore space--time dependence but, as noted in {[}25{]}, without a distribution-free guarantee. I am not aware of work that quantifies day-level overdispersion and clustering of misses across a forecast grid, although my literature search is not exhaustive.

\hypertarget{verification-of-probabilistic-forecasts.}{%
\paragraph{Verification of probabilistic forecasts.}\label{verification-of-probabilistic-forecasts.}}

Meteorological verification uses calibration and sharpness {[}21{]}, rank histograms {[}22{]}, proper scores including the interval score {[}20{]}, and benchmark suites such as WeatherBench Probability {[}16{]}. These tools assess calibration averaged over locations and times; my diagnostics complement them by quantifying day-to-day variation in the fraction of cells covered and the temporal clustering of misses for interval forecasts.

\hypertarget{data-forecaster-and-methods}{%
\section{Data, forecaster and methods}\label{data-forecaster-and-methods}}

\hypertarget{data-and-splits}{%
\subsection{Data and splits}\label{data-and-splits}}

I use ERA5 reanalysis {[}13, 35, 36{]} from the WeatherBench 2 archive {[}15{]} on its \({5.625}^{\circ}\) equiangular grid (H\,=\,32 latitudes, W\,=\,64 longitudes, 2048 cells), at 6-hourly resolution. The target variable is 2\,m temperature (T2m); 500\,hPa geopotential (Z500) is used as an additional forecaster input. The store I used is named as covering 1959--2022 but its time axis ends on 2021-12-31 18~UTC; I use 1979-01-01 onward, 62,824 time steps in total. Forecasts are verified against ERA5 itself, so all coverage statements are relative to the reanalysis, not to independent observations.

Splits are by initialisation date. The forecaster is trained on every second 6-hourly initialisation in 1979--2015 (27,028 initialisations) and model selection uses 2016 at 00~UTC (366 initialisations). Conformal calibration uses 00~UTC initialisations in 2017--2018 (730 initialisations) and evaluation uses 00~UTC initialisations from 2019-01-01 to 2021-12-21 (1,086 consecutive daily initialisations); initialisations after 2021-12-21 are excluded because their 10-day targets lie beyond the data. There is no purge gap between calibration and test: calibration forecasts issued in late December 2018 verify in early January 2019. The lead times are 6\,h, 12\,h, 1, 2, 3, 5, 7 and 10 days. Because every initialisation is at 00~UTC, the 6\,h and 12\,h forecasts verify at 06 and 12~UTC and all other leads at 00~UTC, so for the first two leads lead time and time of day are confounded.

\hypertarget{forecaster}{%
\subsection{Forecaster}\label{forecaster}}

The forecaster is a small residual convolutional network with periodic padding in longitude and replicate padding in latitude (452,744 parameters; a stem convolution, six residual blocks with 64 channels and group normalisation, and a convolutional head). Its seven input channels are the standardised T2m and Z500 fields at the initialisation time, their 6-hour tendencies, the sine and cosine of day of year, and the cosine of latitude. It predicts the T2m change at all eight leads simultaneously, scaled per lead, and is trained with mean-squared error (AdamW, learning rate 2\,×\,10\ensuremath{^{-}}³ with a one-cycle schedule, weight decay 10\ensuremath{^{-}}⁴, batch size 64, 8 epochs, seed 0). I keep the checkpoint with the lowest 2016 validation loss, which was reached at epoch 3 (training loss then 0.402, validation loss 0.571 in normalised units); validation loss rose afterwards while training loss kept falling. Conformal procedures are model-agnostic, so I require only a reasonable forecaster, and I make no attempt to optimise it.

\hypertarget{interval-procedures}{%
\subsection{Interval procedures}\label{interval-procedures}}

Let \({\widehat{y}}_{n,l,i}\) be the forecast issued at initialisation n for lead l at cell i, \(y_{n,l,i}\) the verifying ERA5 value, and \(r_{n,l,i} = \left| y_{n,l,i} - {\widehat{y}}_{n,l,i} \right|\). All intervals have the form \ensuremath{\hat{y}}\,±\,q with half-width q\,≥\,0. For a set of m residuals, \(Q_{1 - \alpha}\) denotes the \ensuremath{\lceil}(m+1)(1−α)\ensuremath{\rceil}-th smallest value (infinite if this exceeds m). The calibration set has \(n_{c} = 730\) initialisations and the test set has \(N_{t} = 1,086\). I use α\,\ensuremath{\in}\,\{0.1,\,0.2,\,0.05\}, with α\,=\,0.1 as the primary level.

\begin{itemize}
\item
  \textbf{Gaussian (RMSE).} \(q_{l,i} = z_{1 - \alpha/2}\,{\widehat{\sigma}}_{l,i}\) with \({\widehat{\sigma}}_{l,i}\) the root-mean-square calibration residual at that lead and cell.
\item
  \textbf{Split, pooled.} One quantile per lead, \(q_{l} = Q_{1 - \alpha}\) of the calibration residuals pooled over all calibration initialisations and all cells.
\item
  \textbf{Split, per-cell.} \(q_{l,i} = Q_{1 - \alpha}\) of the calibration residuals at that lead and cell.
\item
  \textbf{Per-cell + season (Mondrian).} As per-cell split, but computed separately within each meteorological season (DJF, MAM, JJA, SON) of the initialisation date.
\item
  \textbf{ACI per-cell, γ\,\ensuremath{\in}\,\{0.005,\,0.02\}.} At each lead and cell the working level evolves as \(\alpha_{t} = \alpha + \gamma\sum_{s \leq t - d_{l}}^{}\left( \alpha - \mathrm{err}_{s} \right)\), where \(\mathrm{err}_{s} = \mathbf{1}\left\lbrack r_{s} > q_{s} \right\rbrack\) and \(d_{l}\) is the feedback delay below. The half-width is the \(\lceil\left( n_{c} + 1 \right)\left( 1 - \alpha_{t} \right)\rceil\)-th smallest calibration residual with the rank clipped to \(\left\lbrack 1,n_{c} \right\rbrack\), so intervals never exceed the largest calibration residual. The calibration set itself is not updated.
\item
  \textbf{Rolling window per-cell, 90 and 365 days.} The finite-sample quantile of the most recent 90 (or 365) verified residuals at that lead and cell, drawn from the calibration period and from test initialisations already verified.
\end{itemize}

Exchangeability does not hold here (seasonality, serial correlation), so none of these procedures carries a formal finite-sample guarantee on my test data; I treat coverage as an empirical quantity.

\hypertarget{delayed-feedback}{%
\subsection{Delayed feedback}\label{delayed-feedback}}

A forecast at lead τ hours issued at initialisation n can be verified only τ hours later. With daily initialisations, an online method issuing forecast n may use residuals only from initialisations \(n\prime \leq n - d_{l}\), where \(d_{l} = \max\left( 1,\lceil\tau_{l}/24\, h\rceil \right)\). This gives delays of 1, 1, 1, 2, 3, 5, 7 and 10 days for the eight leads. Both ACI and the rolling-window methods obey this rule.

\hypertarget{evaluation-metrics}{%
\subsection{Evaluation metrics}\label{evaluation-metrics}}

Let \(\varphi_{h}\) be the latitude of row h and \(w_{i} = \cos\varphi_{h}/\left( W\sum_{h\prime}^{}\cos\varphi_{h\prime} \right)\) for cell i in row h, so that \(\sum_{i}^{}w_{i} = 1\). Write \(C_{n,l,i} = \mathbf{1}\left\lbrack r_{n,l,i} \leq q_{n,l,i} \right\rbrack\) and \(c_{n,l} = \sum_{i}^{}w_{i}C_{n,l,i}\) for the area-weighted fraction of cells covered by forecast (n,l). \emph{Coverage} is \({\bar{c}}_{l} = N_{t}^{- 1}\sum_{n}^{}c_{n,l}\). \emph{Width} is the area-weighted mean of 2q. The \emph{interval score} {[}20{]} of a symmetric interval is \(2q + \frac{2}{\alpha}(r - q)_{+}\), averaged with area weights; lower is better.

\hypertarget{dependence-diagnostics}{%
\subsection{Dependence diagnostics}\label{dependence-diagnostics}}

\hypertarget{spatial-overdispersion-index.}{%
\paragraph{Spatial overdispersion index.}\label{spatial-overdispersion-index.}}

If cells were covered independently with miss probabilities \(m_{l,i}\), the daily area-weighted coverage \(c_{n,l}\) would have variance \(\sum_{i}^{}w_{i}^{2}\, m_{l,i}\left( 1 - m_{l,i} \right)\). I define

\[\mathrm{VIF}_{l} = \frac{\mathrm{Var}_{n}\left( c_{n,l} \right)}{\sum_{i}^{}w_{i}^{2}\, m_{l,i}\left( 1 - m_{l,i} \right)},\]

with \(m_{l,i}\) the test miss rate of cell i. VIF\,=\,1 means no excess day-to-day variation relative to independent cells with the same miss rates; VIF\,\ensuremath{\gg}\,1 means that on particular days misses cluster. Var denotes the sample variance (n − 1) in the benchmark tables; the day-scale analyses (Sections 4.8 to 4.11 and Appendix G) use the population variance, which differs by a factor n/(n − 1), about 0.1\%. A simpler variant, \(\mathrm{VIF}_{l}^{\mathrm{hom}}\), replaces the denominator by \({\widehat{p}}_{l}\left( 1 - {\widehat{p}}_{l} \right)\sum_{i}^{}w_{i}^{2}\) with \({\widehat{p}}_{l} = 1 - {\bar{c}}_{l}\), i.e.~assumes a common miss rate. The heterogeneity-aware version was adopted after I observed an implausible dip of \(\mathrm{VIF}^{\mathrm{hom}}\) for pooled split conformal prediction at 12\,h (Section~4.4); the two agree to within 0.02 for most per-cell methods and within 0.15 for rolling-90 (differences in Appendix~B). I also report \(N_{\mathrm{eff}} = (\sum_{i}^{}w_{i}^{2})^{- 1}/\mathrm{VIF}^{\mathrm{hom}}\), an effective number of independent cells; the numerator is about 1,660 for this grid and weighting. VIF responds both to spatial correlation of errors and to days on which errors are large almost everywhere; I do not attempt to separate these.

\hypertarget{daily-coverage-tail.}{%
\paragraph{Daily coverage tail.}\label{daily-coverage-tail.}}

The 5th percentile and minimum of \(c_{n,l}\) over test days, and the share of forecasts with \(c_{n,l} < 1 - \alpha - \delta\) for thresholds δ. The primary threshold δ\,=\,0.10 was fixed before running experiments; because it almost never triggers, I also report δ\,=\,0.02 and 0.05.

\hypertarget{miss-persistence.}{%
\paragraph{Miss persistence.}\label{miss-persistence.}}

Let \(\pi_{l}\) be the probability, pooled over cells (unweighted) and consecutive test days, that a cell misses on day n given that it missed on day n\,−\,1. I compare it with \(\pi_{l}^{0}\), the same statistic after randomly permuting the order of test days (mean of ten permutations, the same permutation across cells and leads, so spatial structure within a day is preserved). The permutation retains each cell's miss rate but removes temporal order, so \(\pi_{l} - \pi_{l}^{0}\) measures temporal clustering of misses beyond what unequal miss rates alone imply. It includes slowly varying structure such as season or flow regime; I do not separate these components.

\hypertarget{conditional-coverage.}{%
\paragraph{Conditional coverage.}\label{conditional-coverage.}}

Area-weighted coverage by season of initialisation and by test year, and the largest absolute deviation from nominal across seasons (or years) at each lead.

\hypertarget{uncertainty-quantification-and-validation-of-the-diagnostics}{%
\subsection{Uncertainty quantification and validation of the diagnostics}\label{uncertainty-quantification-and-validation-of-the-diagnostics}}

I report 95\% moving-block bootstrap percentile intervals over test initialisations (blocks of 30 days, 1,000 replicates) and paired block bootstrap intervals for differences between methods; a sensitivity check uses blocks of 10 and 60 days. The bootstrap resamples test days and holds the calibration set and the forecaster fixed, so intervals for the split methods understate total uncertainty. Because many intervals are reported, differences are exploratory and not corrected for multiplicity.

To check that the diagnostics do not flag dependence where none exists, I ran them on synthetic exchangeable, spatially independent Gaussian errors (730 test days). In the rows I inspected (Gaussian and pooled split, all eight leads), coverage was close to 0.90, VIF ranged from 0.88 to 1.06, and persistence matched its shuffled-days null within 0.003. These synthetic numbers are a sanity check, not results.

\hypertarget{day-scale-decomposition}{%
\subsection{Day-scale decomposition}\label{day-scale-decomposition}}

For lead l, initialisation n and cell i, let q(l,i) be the per-cell split-conformal quantile from the calibration rows and u(n,l,i) = r(n,l,i)/q(l,i) the normalised residual. The day-scale g(n,l) is the area-weighted mean of u(n,l,i) over cells, with the weights w(i) defined above. I use the heterogeneity-aware overdispersion index of Section~3.6, which divides the variance over days of the area-weighted coverage by the sum of w(i)² m(i)(1 − m(i)), where m(i) is the miss rate of cell i, using the population variance (the repository's code uses the sample variance, which is larger by a factor n/(n − 1), about 0.1 percent).

\textbf{Oracle.} Normalised residuals are divided by g(n,l) (global), or by the area-weighted mean over each of 16 regions of 8 by 16 cells (4 latitude bands by 4 longitude blocks), recalibrated per cell by split conformal on the calibration days, and the index is recomputed on the test days. The oracle uses the verifying truth and is not a forecasting procedure. Because it forces each day's mean to one, it lowers the index mechanically, so it is always read against the null floor below.

\textbf{Null floor.} The identical pipeline is applied after permuting days independently for every cell and lead, which removes all dependence. On my data the base pipeline gives 0.90 to 1.06 and the oracle pipeline 0.43 to 0.61. The share of excess removed is (VIF base − VIF pipeline) / (VIF base − VIF floor).

\textbf{Split-half.} The day-scale is computed from a random half of the cells and applied to the other half, and the index is computed on the second half only (both directions averaged). This avoids the self-normalisation artifact but adds estimation noise, so its floor is higher (1.5 to 2.0). The base index is computed on the same cells for comparison.

\textbf{Index by distance.} VIF − 1 equals the sum, over great-circle distance bins, of the area-weighted miss covariances between pairs of distinct cells, divided by the same denominator, so the contributions add up exactly. I report it descriptively.

\hypertarget{day-scale-predictability-test}{%
\subsection{Day-scale predictability test}\label{day-scale-predictability-test}}

The method was fixed before it was run on any data. I regress log g(n,l) by ordinary least squares on (i) two annual harmonics of the valid day of year, (ii) log g of the 24 h forecast issued the day before, which is verified by the time forecast n is issued, and (iii) log g of the same-lead forecast whose verification arrives at that time. Predictions for calibration days are out-of-fold (fit on one calibration year, predicted on the other), and the test-day model is fit on all calibration days, excluding the first 10 days for which lagged features do not exist. Scores are divided by the predicted day-scale and calibrated per cell. A test day's own residuals are never used for its own scale. I assume ERA5 verification is available when a forecast is valid.

I report the out-of-sample R² of log g, the overdispersion index, coverage, the 5th percentile of daily coverage, the ratio of mean width to per-cell split, and the area-weighted interval score at α, with the 30-day moving-block bootstrap of Section~3.7 (200 replicates here, against 1,000 for the main tables). Four ablations (season only, fresh only, stale only, fresh plus stale) are reported alongside. This is a diagnostic of what is predictable, not a proposed method.

\hypertarget{second-forecaster}{%
\subsection{Second forecaster}\label{second-forecaster}}

I add ECMWF IFS HRES 2 m temperature from WeatherBench 2 on the same 64 by 32 grid, at the same 00 UTC initialisation dates and split labels as the CNN and verified against the same ERA5 targets, with the same eight procedures, delayed-feedback rule, diagnostics and bootstrap. Grid coordinates match the CNN arrays exactly. HRES is an operational system whose model cycle changed during the calibration and test windows (Section~4.10).

\hypertarget{results}{%
\section{Results}\label{results}}

Unless stated otherwise, results are for α\,=\,0.1 (nominal coverage 0.90) on the 1,086 test initialisations. Tables report the three leads 24\,h, 120\,h and 240\,h; all eight leads appear in the figures. Appendix~E repeats the main tables for α\,=\,0.2 and 0.05.

\hypertarget{forecaster-skill}{%
\subsection{Forecaster skill}\label{forecaster-skill}}

\Needspace{9\baselineskip}\medskip\noindent\textbf{Table 1.} Area-weighted RMSE (K) of 2\,m temperature over the test period.

\nopagebreak
\begin{longtable}[]{@{}
  >{\raggedright\arraybackslash}p{(\columnwidth - 16\tabcolsep) * \real{0.1111}}
  >{\raggedright\arraybackslash}p{(\columnwidth - 16\tabcolsep) * \real{0.1111}}
  >{\raggedright\arraybackslash}p{(\columnwidth - 16\tabcolsep) * \real{0.1111}}
  >{\raggedright\arraybackslash}p{(\columnwidth - 16\tabcolsep) * \real{0.1111}}
  >{\raggedright\arraybackslash}p{(\columnwidth - 16\tabcolsep) * \real{0.1111}}
  >{\raggedright\arraybackslash}p{(\columnwidth - 16\tabcolsep) * \real{0.1111}}
  >{\raggedright\arraybackslash}p{(\columnwidth - 16\tabcolsep) * \real{0.1111}}
  >{\raggedright\arraybackslash}p{(\columnwidth - 16\tabcolsep) * \real{0.1111}}
  >{\raggedright\arraybackslash}p{(\columnwidth - 16\tabcolsep) * \real{0.1111}}@{}}
\toprule\noalign{}
\begin{minipage}[b]{\linewidth}\raggedright
Lead
\end{minipage} & \begin{minipage}[b]{\linewidth}\raggedright
6\,h
\end{minipage} & \begin{minipage}[b]{\linewidth}\raggedright
12\,h
\end{minipage} & \begin{minipage}[b]{\linewidth}\raggedright
1\,d
\end{minipage} & \begin{minipage}[b]{\linewidth}\raggedright
2\,d
\end{minipage} & \begin{minipage}[b]{\linewidth}\raggedright
3\,d
\end{minipage} & \begin{minipage}[b]{\linewidth}\raggedright
5\,d
\end{minipage} & \begin{minipage}[b]{\linewidth}\raggedright
7\,d
\end{minipage} & \begin{minipage}[b]{\linewidth}\raggedright
10\,d
\end{minipage} \\
\midrule\noalign{}
\endfirsthead
\toprule\noalign{}
\begin{minipage}[b]{\linewidth}\raggedright
Lead
\end{minipage} & \begin{minipage}[b]{\linewidth}\raggedright
6\,h
\end{minipage} & \begin{minipage}[b]{\linewidth}\raggedright
12\,h
\end{minipage} & \begin{minipage}[b]{\linewidth}\raggedright
1\,d
\end{minipage} & \begin{minipage}[b]{\linewidth}\raggedright
2\,d
\end{minipage} & \begin{minipage}[b]{\linewidth}\raggedright
3\,d
\end{minipage} & \begin{minipage}[b]{\linewidth}\raggedright
5\,d
\end{minipage} & \begin{minipage}[b]{\linewidth}\raggedright
7\,d
\end{minipage} & \begin{minipage}[b]{\linewidth}\raggedright
10\,d
\end{minipage} \\
\midrule\noalign{}
\endhead
\bottomrule\noalign{}
\endlastfoot
Forecaster (CNN) & 1.99 & 2.32 & 1.28 & 1.81 & 2.13 & 2.44 & 2.57 & 2.67 \\
HRES & 0.56 & 0.67 & 0.69 & 0.77 & 0.88 & 1.24 & 1.75 & 2.42 \\
Persistence & 2.56 & 3.58 & 1.73 & 2.34 & 2.62 & 2.90 & 3.06 & 3.20 \\
Climatology (1979--2015) & 2.48 & 2.49 & 2.45 & 2.45 & 2.45 & 2.45 & 2.45 & 2.45 \\
\end{longtable}

The forecaster beats persistence at every lead (Table~1) and beats the 1979--2015 climatology up to 3 days. At 5 days it is about equal to climatology (2.44 vs 2.45\,K) and at 7 and 10 days it is worse (2.57 and 2.67\,K, roughly 5\% and 9\% above climatology), so intervals at long leads describe a forecaster with essentially no skill beyond about five days. RMSE is not monotone in lead: it is larger at 6\,h and 12\,h than at 1 day; HRES, verified at the same times of day, shows no such bump (below), so the pattern is specific to the CNN at short leads and not a product of verification time alone, although lead and time of day remain confounded at 6 h and 12 h (Section 3.1). The RMSE values are for a \({5.625}^{\circ}\) grid, which smooths small-scale error, and are not comparable to operational numerical weather prediction. The mean forecast error (unweighted over cells) is at most 0.42\,K in absolute value at any lead and is similar in the calibration years and in 2019--2020; 2021 differs by about 0.1\,K at the middle and long leads (Appendix~C). HRES (Section~3.10), computed identically on the same grid, has lower RMSE than the CNN at every lead (0.56 K against 1.99 K at 6 h; 2.42 K against 2.67 K at 10 d) and beats climatology up to 7 days; at 10 days its RMSE is about equal to climatology (2.42 against 2.45 K), so intervals at the longest lead describe a forecast with little skill for HRES as well. Its row is comparable with the other rows of Table~1 but not with published operational scores at native resolution.

\hypertarget{marginal-coverage-and-sharpness}{%
\subsection{Marginal coverage and sharpness}\label{marginal-coverage-and-sharpness}}

\Needspace{9\baselineskip}\medskip\noindent\textbf{Table 2.} Coverage, interval score and $\mathrm{VIF}^{\mathrm{hom}}$ at α\,=\,0.1, with 95\% block-bootstrap intervals over test days (calibration set held fixed). Nominal coverage is 0.90.

\nopagebreak
\begin{longtable}[]{@{}
  >{\raggedright\arraybackslash}p{(\columnwidth - 6\tabcolsep) * \real{0.2500}}
  >{\raggedright\arraybackslash}p{(\columnwidth - 6\tabcolsep) * \real{0.2500}}
  >{\raggedright\arraybackslash}p{(\columnwidth - 6\tabcolsep) * \real{0.2500}}
  >{\raggedright\arraybackslash}p{(\columnwidth - 6\tabcolsep) * \real{0.2500}}@{}}
\toprule\noalign{}
\begin{minipage}[b]{\linewidth}\raggedright
Method
\end{minipage} & \begin{minipage}[b]{\linewidth}\raggedright
Coverage
\end{minipage} & \begin{minipage}[b]{\linewidth}\raggedright
Interval score (K)
\end{minipage} & \begin{minipage}[b]{\linewidth}\raggedright
\(\mathrm{VIF}^{\mathrm{hom}}\)
\end{minipage} \\
\midrule\noalign{}
\endfirsthead
\toprule\noalign{}
\begin{minipage}[b]{\linewidth}\raggedright
Method
\end{minipage} & \begin{minipage}[b]{\linewidth}\raggedright
Coverage
\end{minipage} & \begin{minipage}[b]{\linewidth}\raggedright
Interval score (K)
\end{minipage} & \begin{minipage}[b]{\linewidth}\raggedright
\(\mathrm{VIF}^{\mathrm{hom}}\)
\end{minipage} \\
\midrule\noalign{}
\endhead
\bottomrule\noalign{}
\endlastfoot
\multicolumn{4}{@{}l@{}}{%
\emph{Lead 24\,h}} \\
Gaussian (RMSE) & 0.900 {[}0.897, 0.903{]} & 4.62 {[}4.52, 4.71{]} & 3.85 {[}3.20, 4.40{]} \\
Split, pooled & 0.936 {[}0.932, 0.940{]} & 6.53 {[}6.36, 6.68{]} & 5.74 {[}4.64, 6.56{]} \\
Split, per-cell & 0.897 {[}0.894, 0.900{]} & 4.62 {[}4.52, 4.71{]} & 3.84 {[}3.20, 4.38{]} \\
Per-cell + season & 0.898 {[}0.896, 0.900{]} & 4.42 {[}4.31, 4.51{]} & 3.31 {[}2.98, 3.62{]} \\
ACI per-cell (γ\,=\,0.005) & 0.901 {[}0.898, 0.903{]} & 4.59 {[}4.48, 4.67{]} & 3.59 {[}3.06, 3.95{]} \\
ACI per-cell (γ\,=\,0.02) & 0.900 {[}0.898, 0.903{]} & 4.48 {[}4.37, 4.58{]} & 3.74 {[}3.16, 4.26{]} \\
Rolling 90\,d & 0.894 {[}0.890, 0.897{]} & 4.54 {[}4.42, 4.64{]} & 4.37 {[}3.67, 5.01{]} \\
Rolling 365\,d & 0.901 {[}0.898, 0.904{]} & 4.62 {[}4.52, 4.71{]} & 3.85 {[}3.18, 4.41{]} \\
\multicolumn{4}{@{}l@{}}{%
\emph{Lead 120\,h}} \\
Gaussian (RMSE) & 0.893 {[}0.890, 0.897{]} & 8.41 {[}8.21, 8.61{]} & 6.09 {[}4.91, 7.47{]} \\
Split, pooled & 0.939 {[}0.935, 0.944{]} & 12.75 {[}12.42, 13.07{]} & 7.17 {[}5.73, 8.56{]} \\
Split, per-cell & 0.893 {[}0.889, 0.897{]} & 8.42 {[}8.22, 8.61{]} & 6.04 {[}4.89, 7.29{]} \\
Per-cell + season & 0.893 {[}0.890, 0.895{]} & 7.91 {[}7.69, 8.11{]} & 5.10 {[}4.26, 5.74{]} \\
ACI per-cell (γ\,=\,0.005) & 0.900 {[}0.897, 0.904{]} & 8.37 {[}8.16, 8.57{]} & 6.65 {[}5.47, 7.71{]} \\
ACI per-cell (γ\,=\,0.02) & 0.900 {[}0.896, 0.905{]} & 8.24 {[}7.98, 8.45{]} & 6.98 {[}5.65, 8.11{]} \\
Rolling 90\,d & 0.885 {[}0.880, 0.891{]} & 8.33 {[}8.08, 8.56{]} & 7.88 {[}6.20, 9.36{]} \\
Rolling 365\,d & 0.899 {[}0.895, 0.903{]} & 8.42 {[}8.22, 8.62{]} & 6.60 {[}5.33, 8.09{]} \\
\multicolumn{4}{@{}l@{}}{%
\emph{Lead 240\,h}} \\
Gaussian (RMSE) & 0.894 {[}0.890, 0.898{]} & 9.07 {[}8.87, 9.27{]} & 6.51 {[}5.49, 7.45{]} \\
Split, pooled & 0.941 {[}0.936, 0.945{]} & 14.06 {[}13.68, 14.43{]} & 7.62 {[}6.10, 8.96{]} \\
Split, per-cell & 0.893 {[}0.889, 0.898{]} & 9.08 {[}8.88, 9.27{]} & 6.48 {[}5.51, 7.35{]} \\
Per-cell + season & 0.891 {[}0.889, 0.894{]} & 8.55 {[}8.30, 8.77{]} & 5.30 {[}4.43, 6.03{]} \\
ACI per-cell (γ\,=\,0.005) & 0.900 {[}0.896, 0.905{]} & 9.09 {[}8.86, 9.29{]} & 7.60 {[}6.09, 8.74{]} \\
ACI per-cell (γ\,=\,0.02) & 0.900 {[}0.895, 0.906{]} & 9.12 {[}8.84, 9.38{]} & 8.44 {[}6.25, 10.40{]} \\
Rolling 90\,d & 0.879 {[}0.873, 0.886{]} & 9.15 {[}8.86, 9.42{]} & 10.18 {[}7.30, 12.80{]} \\
Rolling 365\,d & 0.899 {[}0.895, 0.904{]} & 9.10 {[}8.89, 9.30{]} & 7.01 {[}5.89, 8.06{]} \\
\end{longtable}

\includegraphics[width=5.83333in,height=3.88889in]{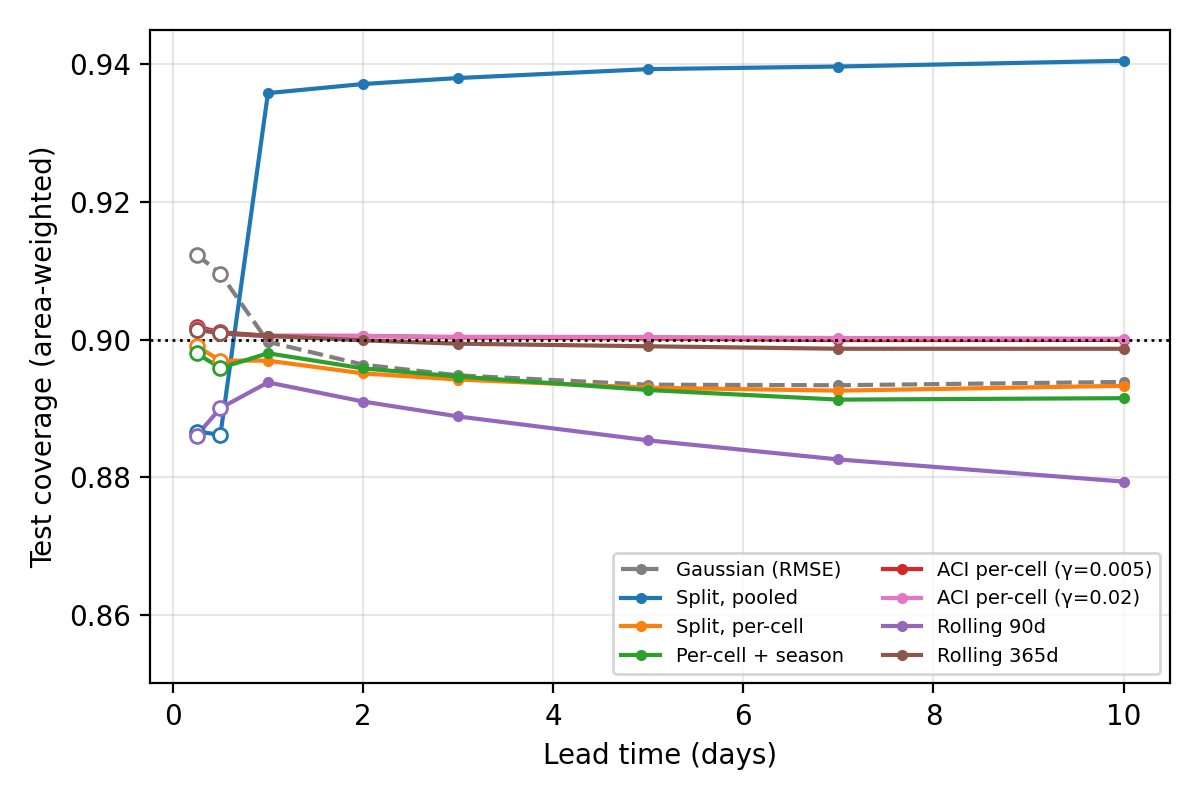}

Figure 1. Area-weighted test coverage by lead time (α\,=\,0.1, nominal 0.90), 2019--2021. Hollow markers: the 6\,h and 12\,h leads, which verify at 06 and 12~UTC while all other leads verify at 00~UTC. Point estimates; 95\% intervals are in Table~2.

Per-cell split conformal prediction, the per-season variant, ACI and the 365-day rolling window all give coverage within about one percentage point of 0.90 at the three leads of Table~2 (Figure~1). The intervals for per-cell split and the per-season variant exclude 0.90 at 120\,h and 240\,h (shortfall of roughly 0.7 to 0.9 points), whereas ACI and the 365-day window include it; because the bootstrap resamples test days only, these intervals are too narrow for the split methods, and I describe the shortfall as small rather than as a failure. The 90-day window drifts below nominal as lead time grows (0.885 at 120\,h, 0.879 at 240\,h). Pooled split conformal prediction over-covers (0.936 to 0.941) and is much wider: its mean interval width is 4.82, 9.74 and 10.79\,K at the three leads, against 3.40, 6.37 and 6.95\,K for per-cell split, with correspondingly worse interval scores (Figure~2).

\includegraphics[width=5.83333in,height=2.33333in]{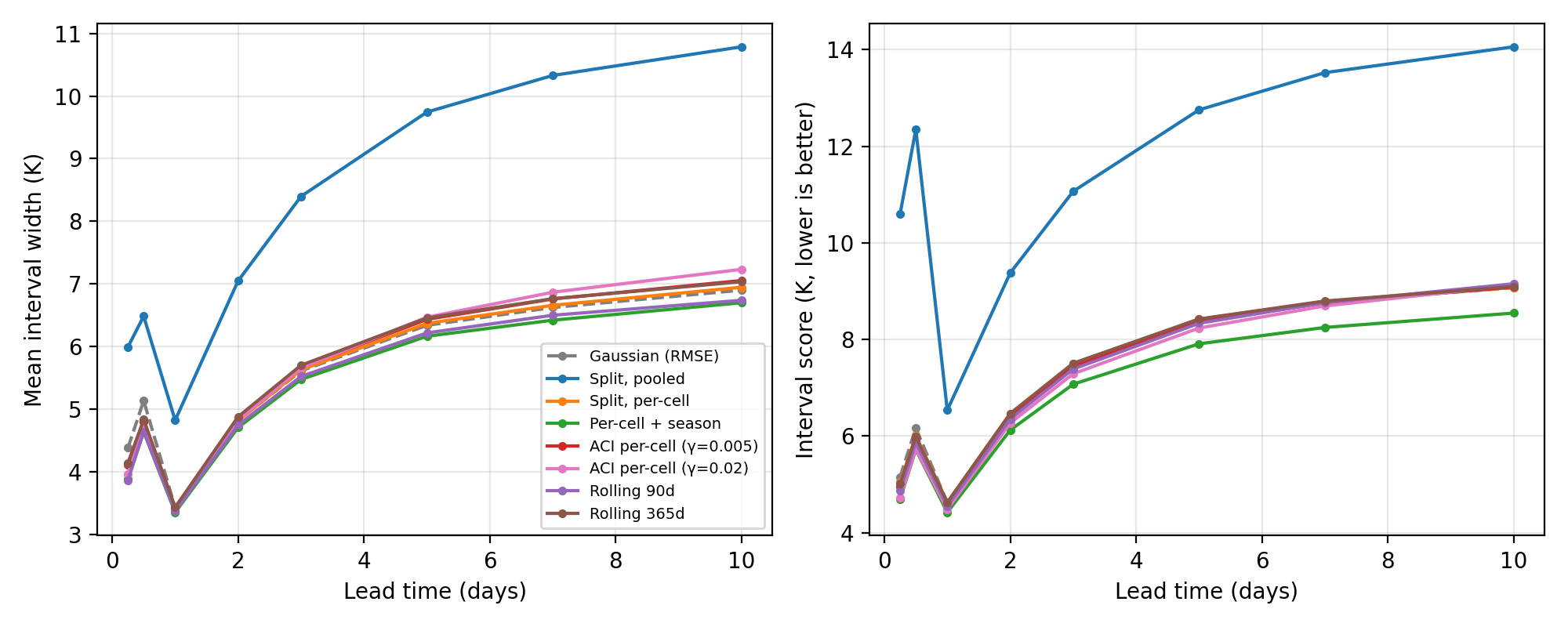}

Figure 2. Mean interval width (left) and interval score (right) by lead time, α\,=\,0.1. Hollow markers as in Figure~1; the bumps at 6 h and 12 h occur at leads where lead and time of day are confounded.

The Gaussian baseline is essentially indistinguishable from per-cell split conformal prediction in interval score: the paired differences are below 0.01\,K at all three leads (Table~5). In this setting the benefit of the conformal methods is the distribution-free construction, not sharper intervals. The per-season variant has the lowest interval score at all three leads, 4.42, 7.91 and 8.55\,K against 4.62, 8.42 and 9.08\,K for per-cell split, an improvement of about 4 to 6 percent.

\hypertarget{spatial-structure-of-coverage}{%
\subsection{Spatial structure of coverage}\label{spatial-structure-of-coverage}}

\includegraphics[width=5.83333in,height=1.81944in]{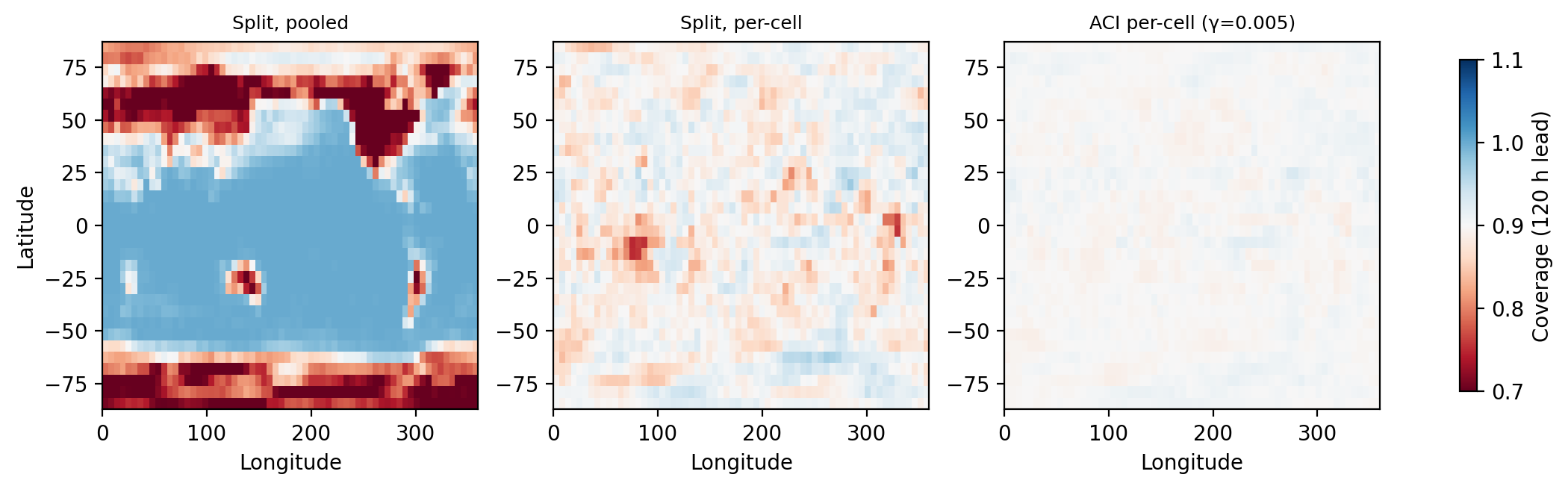}

Figure 3. Test coverage by grid cell at 120\,h lead (α\,=\,0.1) for pooled split, per-cell split and ACI per-cell (γ\,=\,0.005). The colour scale is centred on 0.90.

Figure~3 shows why pooled calibration is poorly suited to a field with strongly heterogeneous error variance. Pooled split over-covers in the tropics and under-covers at mid and high latitudes, with large regions near the lower end of the colour scale (about 0.7). Per-cell split is much more uniform, as it is calibrated cell by cell, but retains a few compact, spatially coherent patches near 0.7 (for example around \(80^{\circ}\)E, 10--\(15^{\circ}\)S and on the equator near 320--\(330^{\circ}\)E); ACI per-cell is nearly uniform at 0.90 in space. A change in mean error between calibration and test does not explain the per-cell patches: across cells, the correlation between the absolute shift in mean error (scaled by the calibration 90th-percentile residual) and the coverage shortfall at 120\,h is about 0.19. I did not identify the cause.

\hypertarget{spatial-overdispersion-and-temporal-clustering}{%
\subsection{Spatial overdispersion and temporal clustering}\label{spatial-overdispersion-and-temporal-clustering}}

\Needspace{9\baselineskip}\medskip\noindent\textbf{Table 3.} Spatial overdispersion index VIF (heterogeneity-aware reference), α\,=\,0.1; point estimates. 1 means no excess day-to-day variation relative to independent cells. Hollow-marker leads (6 h, 12 h) are those at which lead and time of day are confounded.

\nopagebreak
\begin{longtable}[]{@{}
  >{\raggedright\arraybackslash}p{(\columnwidth - 16\tabcolsep) * \real{0.1111}}
  >{\raggedright\arraybackslash}p{(\columnwidth - 16\tabcolsep) * \real{0.1111}}
  >{\raggedright\arraybackslash}p{(\columnwidth - 16\tabcolsep) * \real{0.1111}}
  >{\raggedright\arraybackslash}p{(\columnwidth - 16\tabcolsep) * \real{0.1111}}
  >{\raggedright\arraybackslash}p{(\columnwidth - 16\tabcolsep) * \real{0.1111}}
  >{\raggedright\arraybackslash}p{(\columnwidth - 16\tabcolsep) * \real{0.1111}}
  >{\raggedright\arraybackslash}p{(\columnwidth - 16\tabcolsep) * \real{0.1111}}
  >{\raggedright\arraybackslash}p{(\columnwidth - 16\tabcolsep) * \real{0.1111}}
  >{\raggedright\arraybackslash}p{(\columnwidth - 16\tabcolsep) * \real{0.1111}}@{}}
\toprule\noalign{}
\begin{minipage}[b]{\linewidth}\raggedright
Method
\end{minipage} & \begin{minipage}[b]{\linewidth}\raggedright
6\,h
\end{minipage} & \begin{minipage}[b]{\linewidth}\raggedright
12\,h
\end{minipage} & \begin{minipage}[b]{\linewidth}\raggedright
1\,d
\end{minipage} & \begin{minipage}[b]{\linewidth}\raggedright
2\,d
\end{minipage} & \begin{minipage}[b]{\linewidth}\raggedright
3\,d
\end{minipage} & \begin{minipage}[b]{\linewidth}\raggedright
5\,d
\end{minipage} & \begin{minipage}[b]{\linewidth}\raggedright
7\,d
\end{minipage} & \begin{minipage}[b]{\linewidth}\raggedright
10\,d
\end{minipage} \\
\midrule\noalign{}
\endfirsthead
\toprule\noalign{}
\begin{minipage}[b]{\linewidth}\raggedright
Method
\end{minipage} & \begin{minipage}[b]{\linewidth}\raggedright
6\,h
\end{minipage} & \begin{minipage}[b]{\linewidth}\raggedright
12\,h
\end{minipage} & \begin{minipage}[b]{\linewidth}\raggedright
1\,d
\end{minipage} & \begin{minipage}[b]{\linewidth}\raggedright
2\,d
\end{minipage} & \begin{minipage}[b]{\linewidth}\raggedright
3\,d
\end{minipage} & \begin{minipage}[b]{\linewidth}\raggedright
5\,d
\end{minipage} & \begin{minipage}[b]{\linewidth}\raggedright
7\,d
\end{minipage} & \begin{minipage}[b]{\linewidth}\raggedright
10\,d
\end{minipage} \\
\midrule\noalign{}
\endhead
\bottomrule\noalign{}
\endlastfoot
Gaussian (RMSE) & 3.48 & 3.62 & 3.86 & 4.62 & 5.23 & 6.10 & 6.34 & 6.53 \\
Split, pooled & 8.71 & 3.47 & 9.27 & 11.43 & 11.71 & 12.66 & 12.86 & 13.79 \\
Split, per-cell & 3.70 & 4.40 & 3.85 & 4.58 & 5.23 & 6.03 & 6.24 & 6.47 \\
Per-cell + season & 4.60 & 3.78 & 3.31 & 3.87 & 4.67 & 5.09 & 5.38 & 5.30 \\
ACI per-cell (γ\,=\,0.005) & 3.20 & 3.87 & 3.59 & 4.72 & 5.52 & 6.65 & 7.16 & 7.59 \\
ACI per-cell (γ\,=\,0.02) & 3.45 & 3.48 & 3.74 & 5.23 & 6.06 & 6.97 & 7.98 & 8.43 \\
Rolling 90\,d & 4.28 & 4.36 & 4.40 & 5.90 & 6.71 & 7.98 & 9.19 & 10.33 \\
Rolling 365\,d & 3.67 & 4.39 & 3.85 & 4.90 & 5.66 & 6.60 & 6.74 & 7.00 \\
\end{longtable}

\Needspace{9\baselineskip}\medskip\noindent\textbf{Table 4.} Miss persistence π (probability of a miss given a miss on the previous day), its shuffled-days null π\ensuremath{^{0}}, and the 5th percentile of daily spatial coverage (target 0.90), α\,=\,0.1. The nominal value of π under independence and equal miss rates is 0.10.

\nopagebreak
\begin{center}
\small
\begin{tabular}{lcccccc}
\toprule
 & \multicolumn{2}{c}{24\,h} & \multicolumn{2}{c}{120\,h} & \multicolumn{2}{c}{240\,h} \\
\cmidrule(lr){2-3}\cmidrule(lr){4-5}\cmidrule(lr){6-7}
Method & π\,/\,π\ensuremath{^{0}} & 5th pct & π\,/\,π\ensuremath{^{0}} & 5th pct & π\,/\,π\ensuremath{^{0}} & 5th pct \\
\midrule
Gaussian (RMSE) & 0.201 / 0.103 & 0.874 & 0.393 / 0.111 & 0.861 & 0.438 / 0.111 & 0.862 \\
Split, pooled & 0.329 / 0.231 & 0.911 & 0.526 / 0.245 & 0.913 & 0.584 / 0.244 & 0.916 \\
Split, per-cell & 0.204 / 0.106 & 0.871 & 0.390 / 0.111 & 0.861 & 0.435 / 0.111 & 0.861 \\
Per-cell + season & 0.182 / 0.105 & 0.875 & 0.369 / 0.112 & 0.864 & 0.419 / 0.114 & 0.861 \\
ACI (γ\,=\,0.005) & 0.187 / 0.099 & 0.876 & 0.380 / 0.100 & 0.868 & 0.431 / 0.099 & 0.864 \\
ACI (γ\,=\,0.02) & 0.157 / 0.099 & 0.875 & 0.375 / 0.099 & 0.866 & 0.432 / 0.099 & 0.862 \\
Rolling 90\,d & 0.195 / 0.108 & 0.868 & 0.409 / 0.121 & 0.846 & 0.470 / 0.128 & 0.833 \\
Rolling 365\,d & 0.195 / 0.099 & 0.875 & 0.384 / 0.101 & 0.867 & 0.433 / 0.100 & 0.866 \\
\bottomrule
\end{tabular}
\end{center}

\includegraphics[width=5.83333in,height=1.66667in]{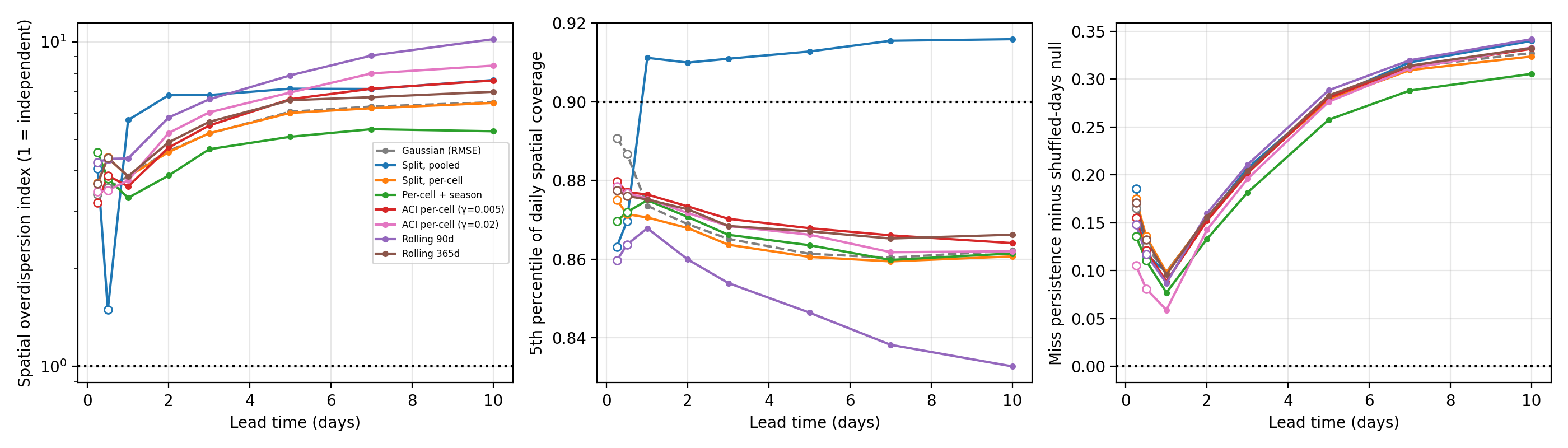}

Figure 4. Dependence diagnostics by lead time, α\,=\,0.1. Left: spatial overdispersion index (1 = independent). Middle: 5th percentile of daily spatial coverage (target 0.90). Right: miss persistence minus its shuffled-days null. Hollow markers: 6\,h and 12\,h leads (see Figure~1). Point estimates.

\hypertarget{overdispersion.}{%
\paragraph{Overdispersion.}\label{overdispersion.}}

For every method the day-to-day variance of the fraction of cells covered is well above the independent-cell value (Table~3, Figure~4). For the per-cell methods VIF is between 3.2 and 4.6 at the shortest leads and rises with lead time to between 5.3 and 10.3 at 10 days; the synthetic control gave 0.88 to 1.06. For per-cell split conformal prediction, \(N_{\mathrm{eff}}\) is 432, 275 and 256 at 24\,h, 120\,h and 240\,h (314 to 503 for the per-season variant), against a reference of about 1,660 independent cells, so the grid behaves as though it had a few hundred independent cells. Pooled split conformal prediction has the largest overdispersion (about 9 to 14 at all leads except 12\,h) but also over-covers, and I do not treat it as a competitive baseline. It has an unexplained dip at 12\,h (3.47 against 8.71 at 6\,h and 9.27 at 24\,h) that survives the heterogeneity-aware reference; I did not identify the cause.

\hypertarget{daily-coverage-tail-and-systemic-failure.}{%
\paragraph{Daily coverage tail and ``systemic failure''.}\label{daily-coverage-tail-and-systemic-failure.}}

The 5th percentile of daily spatial coverage is about 0.86 to 0.88 for the per-cell methods, except rolling-90 at long leads (0.846 and 0.833 at 120\,h and 240\,h; Table~4), and the worst single day among the 1,086 test days lies between 0.78 (rolling-90 at 240 h) and 0.86 (ACI with γ = 0.005 at 24 h) for the non-pooled methods at the three leads of Table 4: on one day in twenty, a nominal 90\% interval covers 86 to 88\% of the globe or less. The pre-specified systemic-failure threshold (δ\,=\,0.10, coverage below 0.80) is essentially never reached: the share of forecasts is 0 for every method except rolling-90 at 168 h and 240 h (0.2\% and 0.4\%). For per-cell split conformal prediction, the share of forecasts with daily coverage below 0.88 (δ\,=\,0.02) is 0.10, 0.15, 0.13, 0.18, 0.21, 0.23, 0.25 and 0.27 at the eight leads (6\,h to 10\,d), the share below 0.85 (δ\,=\,0.05) is at most 0.02, and no forecast falls below 0.80. So in this setting day-to-day failures are modest in size but common.

\hypertarget{temporal-clustering-of-misses.}{%
\paragraph{Temporal clustering of misses.}\label{temporal-clustering-of-misses.}}

The probability of a miss given a miss on the previous day is far above both nominal 0.10 and the shuffled-days null (Table~4). The excess π\,−\,π\ensuremath{^{0}} is 0.06 to 0.10 at 24\,h, 0.26 to 0.29 at 120\,h and 0.31 to 0.34 at 240\,h across all methods (Figure~4, right); the per-season variant is lowest at 120\,h and 240\,h. ACI equalises miss rates across cells (its null is close to 0.10) but its excess at long leads is as large as for the non-adaptive methods. A plausible reading is that at long leads the forecast is close to climatology, so a miss reflects a temperature anomaly, and anomalies persist for days; this is a hypothesis, not a tested explanation. Because the shuffle removes all temporal order, the excess also contains slowly varying components such as season or regime.

\hypertarget{coverage-by-season-and-year}{%
\subsection{Coverage by season and year}\label{coverage-by-season-and-year}}

\includegraphics[width=5.83333in,height=2.12121in]{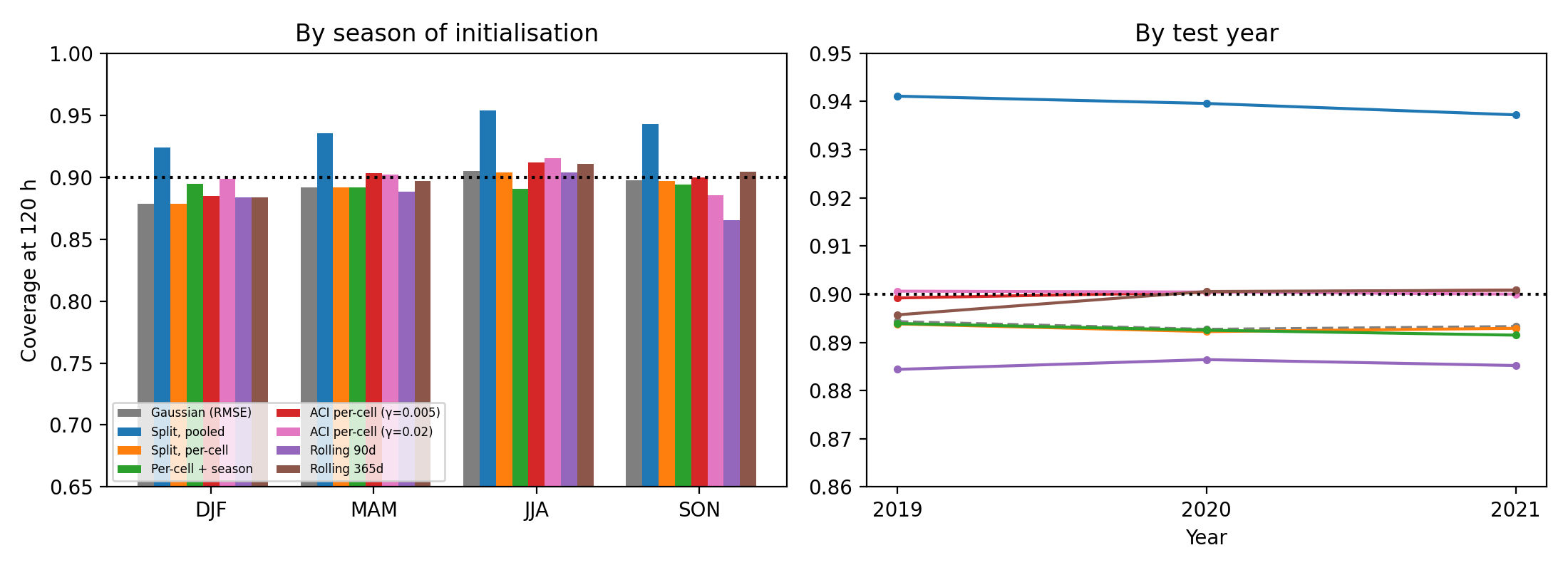}

Figure 5. Coverage at 120\,h by season of initialisation (left) and by test year (right), α\,=\,0.1.

Season matters more than year (Figure~5). The largest absolute deviation from nominal across seasons at 24, 120 and 240\,h is 0.011, 0.021 and 0.021 for per-cell split, 0.004, 0.010 and 0.013 for the per-season variant, 0.010 to 0.018 for ACI and 0.019, 0.035 and 0.047 for the 90-day window. Across years the largest deviation is at most 0.009 for the per-cell, Gaussian, ACI and 365-day methods, 0.022 for the 90-day window at 240\,h and 0.036 to 0.042 for pooled split. I see no monotone drift across 2019--2021 in this window. The gap between calibration and test years therefore did not produce a visible drift-driven loss of coverage here, and adaptivity gains little on this evidence.

\hypertarget{paired-comparisons-between-methods}{%
\subsection{Paired comparisons between methods}\label{paired-comparisons-between-methods}}

\Needspace{9\baselineskip}\medskip\noindent\textbf{Table 5.} Paired differences with split per-cell conformal prediction as reference, α\,=\,0.1; 95\% paired block-bootstrap intervals over test days. Negative score differences favour the method named. “–”: not computed.

\nopagebreak
\begin{longtable}[]{@{}
  >{\raggedright\arraybackslash}p{(\columnwidth - 8\tabcolsep) * \real{0.2000}}
  >{\raggedright\arraybackslash}p{(\columnwidth - 8\tabcolsep) * \real{0.2000}}
  >{\raggedright\arraybackslash}p{(\columnwidth - 8\tabcolsep) * \real{0.2000}}
  >{\raggedright\arraybackslash}p{(\columnwidth - 8\tabcolsep) * \real{0.2000}}
  >{\raggedright\arraybackslash}p{(\columnwidth - 8\tabcolsep) * \real{0.2000}}@{}}
\toprule\noalign{}
\begin{minipage}[b]{\linewidth}\raggedright
Contrast
\end{minipage} & \begin{minipage}[b]{\linewidth}\raggedright
Lead
\end{minipage} & \begin{minipage}[b]{\linewidth}\raggedright
Δ interval score (K)
\end{minipage} & \begin{minipage}[b]{\linewidth}\raggedright
\(\Delta\mathrm{VIF}^{\mathrm{hom}}\)
\end{minipage} & \begin{minipage}[b]{\linewidth}\raggedright
Δ coverage
\end{minipage} \\
\midrule\noalign{}
\endfirsthead
\toprule\noalign{}
\begin{minipage}[b]{\linewidth}\raggedright
Contrast
\end{minipage} & \begin{minipage}[b]{\linewidth}\raggedright
Lead
\end{minipage} & \begin{minipage}[b]{\linewidth}\raggedright
Δ interval score (K)
\end{minipage} & \begin{minipage}[b]{\linewidth}\raggedright
\(\Delta\mathrm{VIF}^{\mathrm{hom}}\)
\end{minipage} & \begin{minipage}[b]{\linewidth}\raggedright
Δ coverage
\end{minipage} \\
\midrule\noalign{}
\endhead
\bottomrule\noalign{}
\endlastfoot
Per-cell + season & 24\,h & {[}−0.25,−0.15{]} & {[}−1.04,0.08{]} & {[}−0.002,0.004{]} \\
& 120\,h & {[}−0.65,−0.36{]} & {[}−2.15,0.06{]} & {[}−0.004,0.003{]} \\
& 240\,h & {[}−0.67,−0.39{]} & {[}−2.36,−0.13{]} & {[}−0.006,0.002{]} \\
ACI (γ\,=\,0.005) & 24\,h & {[}−0.07,−0.01{]} & {[}−0.74,0.13{]} & {[}0.002,0.005{]} \\
& 120\,h & {[}−0.12,0.02{]} & {[}−0.15,1.31{]} & {[}0.005,0.009{]} \\
& 240\,h & {[}−0.06,0.08{]} & {[}0.19,1.91{]} & {[}0.005,0.009{]} \\
Rolling 90\,d & 24\,h & {[}−0.14,−0.02{]} & {[}−0.25,1.34{]} & {[}−0.007,0.000{]} \\
& 120\,h & {[}−0.26,0.09{]} & {[}−0.22,3.74{]} & {[}−0.013,−0.003{]} \\
& 240\,h & {[}−0.12,0.27{]} & {[}0.90,6.38{]} & {[}−0.020,−0.008{]} \\
Gaussian (RMSE) & 24\,h & {[}−0.005,0.000{]} & -- & -- \\
& 120\,h & {[}−0.007,0.000{]} & -- & -- \\
& 240\,h & {[}−0.009,0.000{]} & -- & -- \\
\end{longtable}

Table~5 supports three statements. First, the per-season variant improves interval score over per-cell split at all three leads and lowers VIF at 240\,h; its VIF intervals at the other two leads just include zero. Second, ACI moves coverage closer to nominal by about 0.5 to 0.9 points at 120\,h and 240\,h without a detectable change in interval score; its VIF at 240\,h is higher than for per-cell split, but this is weak evidence: the lower end of the interval approaches zero with longer blocks (paired intervals {[}0.43, 1.78{]}, {[}0.19, 1.91{]} and {[}0.06, 1.92{]} for blocks of 10, 30 and 60 days), and the difference is not distinguishable from zero at α\,=\,0.2 (Appendix~E). Third, the 90-day window has lower coverage and higher overdispersion at long leads. I report the ACI comparison only as an exploratory observation; I did not test a mechanism such as the 10-day feedback delay.

\hypertarget{other-significance-levels}{%
\subsection{Other significance levels}\label{other-significance-levels}}

At α\,=\,0.2 and 0.05 the qualitative picture is unchanged (Appendix~E): per-cell split and the per-season variant are slightly below nominal while ACI and the 365-day window are close to it, pooled split over-covers, VIF is well above 1 and grows with lead, the persistence excess grows from about 0.04 to 0.10 at 24\,h to 0.25 to 0.38 at 240\,h, and the per-season variant has a lower interval score than per-cell split at the three leads examined, with paired intervals excluding zero. The one exception is the ACI overdispersion difference at 240\,h noted above.

\hypertarget{a-day-scale-explains-most-of-the-excess}{%
\subsection{A day-scale explains most of the excess}\label{a-day-scale-explains-most-of-the-excess}}

Table~12 (Appendix~G) reports the decomposition for the CNN at α = 0.1. The global oracle brings the overdispersion index from 3.7 to 6.5 down to 0.8 to 1.3, against a null floor of 0.5 to 0.6 for the same pipeline. Measured from the base value to that floor, the oracle removes 81 to 92 percent of the excess with one global factor per day and 87 to 96 percent with 16 regional factors. The oracle intervals stay above the floor at every lead (lower bounds 0.70 to 1.10 against floors of 0.48 to 0.61; the floors are single estimates), so some dependence remains beyond a global day-scale, by about 0.7 at 5 to 10 days.

Because the oracle is biased low, I also use the split-half version. It lowers the index at every lead from 12 h to 10 d, with the bootstrap interval for the drop above zero (lower bounds 0.32 to 1.30); at 6 h the interval includes zero. Its floor is high (1.7 to 2.0), so it has limited resolution.

Splitting VIF − 1 by the distance between cell pairs shows large positive contributions from pairs 1,000 to 8,000 km apart and strongly negative contributions (about −6 to −8 at leads of one day and longer) from pairs 12,000 to 20,100 km apart. The net index therefore hides large opposing terms, and I do not interpret the negative long-distance terms.

\hypertarget{predicting-the-day-scale}{%
\subsection{Predicting the day-scale}\label{predicting-the-day-scale}}

Table~13 reports the day-scale model of Section~3.9 for both forecasters. For the CNN, the out-of-sample R² of log g is 0.31 to 0.51 and the index falls by 18 to 44 percent, with coverage within 0.01 of nominal (0.892 to 0.899) and a width ratio of 0.99 to 1.00 relative to per-cell split. The drop is distinguishable from zero at every lead and every block length tried (30, 60 and 90 days; tightest lower bound 0.15, at 6 h with 90-day blocks). The interval score improves by only 0.1 to 0.6 percent, with the interval for the difference excluding zero at seven of eight leads, and the 5th percentile of daily coverage barely moves (0.871 to 0.879 at 1 d, 0.861 to 0.864 at 10 d).

At 5 to 10 days the seasonal term alone gives about the same index as the full model (4.67, 4.67 and 4.51 against 4.65, 4.70 and 4.67), so what is predictable there is mainly seasonal. Verified-error predictors matter at short leads.

\hypertarget{a-second-forecaster-hres}{%
\subsection{A second forecaster: HRES}\label{a-second-forecaster-hres}}

HRES, an operational numerical forecast, has lower area-weighted RMSE than the CNN at every lead (Table~1), from 0.56 K against 1.99 K at 6 h to 2.42 K against 2.67 K at 10 d, where it is about equal to climatology. Table~14 gives the decomposition. The overdispersion is not an artifact of a weak forecaster: the base index is 3.1 to 9.0, and at 3, 5 and 7 days its lower bounds are 2.72, 3.67 and 5.17. It is U-shaped in lead, highest at 6 to 12 h and at 10 d and lowest at 3 d, and the 6 to 12 h intervals are very wide (2.7 to 14.1 and 3.6 to 11.7). The global oracle removes 93 to 98 percent of the excess, relative to its floor of 0.43 to 0.54. The split-half drop excludes zero at all leads but 3 d (−0.07 to 0.34), where the evidence for a global day-scale rests mainly on the oracle.

HRES is not a fixed system, and per-cell split coverage drifts across the test years at short leads (0.886, 0.879 and 0.864 in 2019, 2020 and 2021 at 6 h), while the CNN's is flat at every lead (0.892 to 0.900). The calibration window spans IFS cycles 43r3 (operational from July 2017) and 45r1 (5 June 2018); the test window spans 46r1 (11 June 2019), 47r1 (30 June 2020), 47r2 (11 May 2021) and 47r3 (12 October 2021) {[}30--34{]}. Coverage is above nominal at 5 to 7 days (0.903 to 0.910). Removing each year's mean daily coverage lowers the base index from 9.01 to 7.63 at 6 h (about 15 percent) and by 6 percent or less at every other lead (Table~17), so year-to-year drift does not drive the overdispersion. This is a lower bound on the contribution of drift, because it removes year means only.

The day-scale model lowers the HRES index by 7 to 66 percent (Table~13). With 1,000 bootstrap replicates and 30-day blocks the interval for the drop is above zero at seven of eight leads and includes zero at 10 d (−0.20 to 1.42); the same holds with 60- and 90-day blocks (200 replicates; tightest lower bound 0.09, at 5 d with 90-day blocks). Its out-of-sample R² is 0.74 at 12 h and falls below 0.2 from 3 d on, and the verified-error predictors add nothing beyond the seasonal term from 5 d. Coverage rises at short leads (0.876 to 0.887 at 6 h) with widths 2 to 3 percent larger, so part of the short-lead gain is a correction of under-coverage that I have not separated from flow dependence. The interval score improves by 0.1 to 0.4 percent, with the interval for the improvement excluding zero from 6 h to 3 d, touching zero at 5 d (0.000 to 0.009) and including zero at 7 and 10 d, and the 5th percentile of daily coverage rises from 0.809 to 0.860 at 6 h and from 0.839 to 0.875 at 12 h. The high short-lead predictability may partly reflect a persistent mismatch between the HRES analysis and ERA5, which I did not test.

\hypertarget{benchmark-methods-on-hres}{%
\subsection{Benchmark methods on HRES}\label{benchmark-methods-on-hres}}

I ran all eight procedures of Section~3.3 on HRES with the same splits, delayed feedback and diagnostics. Tables~15 and 16 give the index and interval score at α = 0.1, with the day-scale model for comparison.

Differences in the index between methods are not established: the unpaired intervals overlap heavily (for the homogeneous VIF at 1 d, 3.79 {[}2.90, 4.37{]} for per-cell split and 3.38 {[}2.45, 4.30{]} for ACI with γ = 0.02). Adaptive methods help at short leads. ACI (γ = 0.02) improves the interval score by about 7 percent at 6 to 24 h and lowers the index at 6 and 12 h (9.02 to 3.91 and 8.05 to 4.91), consistent with slowly varying error levels that online methods can track. From 1 d onward they change it little. Season-conditional calibration improves the interval score by 4 to 5 percent at 3 to 10 d and has the lowest index among the eight procedures at 2 to 10 d (2.86 to 7.38), but it stays well above 1. At the three leads with intervals (1, 5 and 10 d), the 5th percentile of daily coverage is 0.85 to 0.89 for every method except pooled split conformal, which reaches 0.91 to 0.93 only by over-covering.

Paired comparisons with the day-scale model (Tables~18 and 19; the same resampled days for both methods, 1,000 replicates, 30-day blocks) show a trade-off. From 12 h to 3 d the day-scale model has a lower index than every benchmark method, with paired intervals excluding zero. At 6 h it is indistinguishable from ACI (γ = 0.02) and from rolling-90, and from 5 to 10 d it is indistinguishable from season-conditional calibration and, at most of those leads, from ACI (γ = 0.02). On interval score the benchmark methods are better. ACI (γ = 0.02) is ahead by 0.13 to 0.14 K from 6 h to 1 d and by 0.06 to 0.11 K from 2 to 5 d, and season-conditional calibration is ahead by 0.05 K at 6 h, rising to 0.41 K at 10 d, with paired intervals excluding zero at every lead. The day-scale model improves on per-cell split conformal by only 0.004 to 0.010 K. At 6 h ACI (γ = 0.02) is therefore at least as good as the day-scale model on both metrics; from 12 h to 3 d no method is best on both; and from 5 to 10 d season-conditional calibration is at least as good on both.

\hypertarget{discussion}{%
\section{Discussion}\label{discussion}}

\hypertarget{what-holds-and-what-does-not.}{%
\paragraph{What holds and what does not.}\label{what-holds-and-what-does-not.}}

In this setting the standard procedures behave well in the sense their guarantees address. Per-cell conformal methods give close to nominal coverage averaged over days, pooled calibration over-covers by pooling cells with very different error scales, and I see no clear drift over three test years. What the marginal view hides is how coverage varies from day to day and across days. The fraction of cells covered on a given day varies several times more than independent cells would allow, a nominal 90\% interval covers 86 to 88\% of the area or less on one day in twenty, and misses cluster in time well beyond what unequal per-cell miss rates imply. Both effects grow with lead time and are present for every method, including those that hit 0.90 on average. The overall outcome is therefore mixed rather than a failure: marginal validity is largely intact, but day-level reliability is weaker than the nominal level suggests.

\hypertarget{why-the-failure-mode-looks-like-this.}{%
\paragraph{Why the failure mode looks like this.}\label{why-the-failure-mode-looks-like-this.}}

I did not isolate causes, but the results are consistent with errors whose magnitude and sign are coherent across space and persist over days. Three observations point in that direction: the overdispersion and persistence excess are shared by methods with very different calibration schemes; they grow with lead as the forecaster approaches climatology; and a per-season quantile, which tracks the seasonal part of error scale, removes only a modest share of either. Online adaptation does not remove them: ACI and rolling windows update each cell separately from stale outcomes, whereas day-to-day swings in coverage involve many cells at once and, at 10 days, the feedback is 10 days old. My dependence diagnostics cannot distinguish spatially correlated errors from days of uniformly large errors, and the decomposition in Section~4.8 shows only that a broad day-scale accounts for most of the excess, not which of the two it is.

\hypertarget{practical-reading.}{%
\paragraph{Practical reading.}\label{practical-reading.}}

A user who needs only long-run calibration at a location can rely on per-cell split or per-season conformal prediction here; the per-season variant has the best interval score. A user who needs day-level or region-level reliability should not read a 90\% interval as a 90\% coverage statement for a single forecast field. The Gaussian baseline is competitive in this setting, so the case for conformal prediction rests on its distribution-free construction rather than on sharper intervals.

\hypertarget{what-the-day-scale-means.}{%
\paragraph{What the day-scale means.}\label{what-the-day-scale-means.}}

The oracle shows that days on which normalised errors are large across much of the grid account for most of the clustering of misses. My diagnostics cannot tell uniformly large errors from large-scale error patterns that cover much of the globe, because both appear as a broad day-scale at these distances.

\hypertarget{a-trade-off-not-a-fix.}{%
\paragraph{A trade-off, not a fix.}\label{a-trade-off-not-a-fix.}}

Predicting the day-scale from information verified at issuance lowers the index without widening intervals, but the predictable part is mainly seasonal and short-lead persistence, and the gain in interval score is at most 0.6 percent. Existing methods do better on interval score (ACI at short leads on HRES, season-conditional calibration at 3 to 10 days) and do not remove the clustering. I therefore treat the day-scale model as a diagnostic of what is predictable, not as a competing method. A dependence-aware method has to beat ACI and season-conditional calibration on interval score and the day-scale model on the index at once. The oracle gap suggests more is available, but it needs information about the day's error size that past verified errors do not supply beyond a few days.

\hypertarget{a-skilful-second-forecaster.}{%
\paragraph{A skilful second forecaster.}\label{a-skilful-second-forecaster.}}

HRES shows the same overdispersion as the CNN despite higher skill at all but the longest leads, so the effect is not an artifact of a weak forecaster. Its short-lead behaviour differs: coverage drifts across years, and adaptive methods reduce the index at 6 to 12 h. For HRES, exchangeability between calibration and test fails by construction, so results at 6 to 24 h may mix dependence with drift.

\hypertarget{limitations}{%
\section{Limitations}\label{limitations}}

\begin{itemize}
\item
  \textbf{Reanalysis target.} Coverage is relative to ERA5, which is itself model output; results may differ against station or satellite observations.
\item
  \textbf{Modest forecaster.} The forecaster has no skill beyond about five days and was not tuned. Conclusions about dependence of errors may change for stronger models or for ensemble forecasts, whose error structure differs. I add a skilful second forecaster, ECMWF HRES, in Sections~4.10 and 4.11; the dependence effects replicate there.
\item
  \textbf{Scope.} One variable (2\,m temperature), one coarse resolution (\({5.625}^{\circ}\)), 00~UTC initialisations only, and a lead time confounded with time of day at 6\,h and 12\,h. Results may not transfer to other variables, finer grids or regions.
\item
  \textbf{Short windows.} Calibration is two years and test is three; calibration residuals are serially correlated, so the effective calibration sample is smaller than 730, and the finite-sample guarantee does not strictly apply. There is no purge gap between calibration and test.
\item
  \textbf{Uncertainty quantification.} The bootstrap holds the calibration set and the forecaster fixed and resamples test days only, so intervals understate total uncertainty. Many comparisons are reported without multiplicity correction; the ACI overdispersion contrast in particular is weak and not robust across α.
\item
  \textbf{Post hoc choices.} The heterogeneity-aware overdispersion reference was adopted after I saw the pooled-split dip, and the additional thresholds δ\,=\,0.02,\,0.05 were added after the pre-specified δ\,=\,0.10 proved uninformative; I report both versions.
\item
  \textbf{Diagnostics are descriptive.} VIF and the persistence excess do not separate spatial correlation from days of uniformly large errors, or temporal autocorrelation from slow regime or seasonal variation. They quantify the symptoms, not the causes. Sections~4.8 and 4.9 add a decomposition and a predictability test, but they still cannot separate uniformly large errors from large-scale error patterns.
\item
  \textbf{Unexplained features.} The 12\,h dip in pooled-split overdispersion and the spatially coherent under-covered patches of per-cell split are not explained.
\item
  \textbf{Day-scale diagnostics.} The day-scale is a single global factor per day and lead; regional factors are used only in the oracle. The oracle uses the verifying truth and is biased low by self-normalisation. I control this with a null floor and a split-half variant, whose floor is high (1.5 to 2.0), so its resolution is limited; at 3 d on HRES its interval includes zero. The negative long-distance terms in the decomposition by distance are unexplained.
\item
  \textbf{HRES is not a fixed system.} ECMWF changed the operational cycle several times across the calibration and test windows, so exchangeability between calibration and test fails for HRES in a way it does not for the CNN. Coverage drifts across test years at short leads, and HRES forecasts verified against ERA5 include an analysis mismatch that the CNN does not have, which matters most at short leads.
\item
  \textbf{Predictability test.} It assumes ERA5 verification is available when a forecast is valid, whereas real-time analyses arrive with a delay. Calibration uses 720 out-of-fold days, and interval-score gains are very small.
\item
  \textbf{Analysis path.} Analyses were explored on the CNN before HRES was run. The day-scale predictability method was fixed before its first run on any data, and the decision rules for HRES were fixed before the HRES runs; all variants I ran are reported. Bootstrap intervals use 30-day blocks (200 replicates for the CNN decomposition table; 1,000 for the HRES decomposition and the day-scale and paired HRES comparisons; 60- and 90-day blocks were also run for the day-scale model on both forecasters), and with about three test years the 10-day intervals are rough.
\end{itemize}

\hypertarget{future-work-a-dependence-aware-method}{%
\section{Future work: a dependence-aware method}\label{future-work-a-dependence-aware-method}}

The failure mode observed here is not drift and not a gross marginal miscalibration; it is dispersion of the daily fraction of cells covered and clustering of misses in time, increasing with lead. This motivates methods that treat the forecast field, not the individual cell, as the unit of calibration. The following directions are proposals motivated by my results; only the first has been tested, in a minimal global form (Sections~4.8 to 4.11).

\begin{enumerate}
\def\labelenumi{\arabic{enumi}.}
\item
  \textbf{Day-adaptive error scale.} Fixed per-cell, per-season quantiles cannot track flow-dependent error magnitude. Normalising scores by a scale estimate \({\widehat{s}}_{n,l,i}\) that combines the climatological per-cell scale with a day-level factor estimated from information available at issuance (for example recent verified residual fields at short leads, or ensemble spread where available) and calibrating the normalised scores within regions could reduce the dispersion of daily coverage at fixed marginal coverage. In Section~4.9 I test a minimal version, a single global day-scale predicted from seasonal terms and already-verified errors. It lowers the overdispersion index without widening intervals but improves interval score by at most 0.6 percent and does not match ACI or season-conditional calibration on interval score; a successful method needs predictors of the day's error size that verified past errors do not provide, such as ensemble spread.
\item
  \textbf{Spatially coupled online updates.} ACI per cell ignores that misses are coherent in space. Updating the working level at a cell from region-smoothed verified outcomes would pool information across neighbours; at long leads the feedback delay still limits what can be learned, so this should be paired with (1).
\item
  \textbf{Calibrating the tail of daily coverage.} Instead of the marginal miss rate, calibrate a region-level score (for example a high quantile of normalised residuals within a region) so that the probability that the fraction of region cells covered falls below a threshold is controlled. This targets the 5th-percentile statistic reported here.
\item
  \textbf{Regime-conditional calibration.} Weighting calibration days by similarity of the large-scale state (analogue days), or by recency, in the spirit of weighted conformal prediction {[}7{]}, could address the temporal clustering of misses.
\end{enumerate}

I would evaluate any such method with the protocol of this paper, fixed in advance: the same splits, delayed feedback for online procedures, and the diagnostics VIF, 5th percentile of daily coverage, systemic-failure share at fixed δ and persistence excess, with paired block-bootstrap comparisons against per-cell and per-season split conformal prediction at matched marginal coverage and equal or better interval score. Further extensions are other variables (for example Z500 and precipitation), finer grids, longer calibration windows, several forecasters including ensemble forecasts, and verification against observations. Given Section~4.11, such a method has to beat ACI and season-conditional calibration on interval score and the day-scale model on the index at once.

\hypertarget{conclusion}{%
\section{Conclusion}\label{conclusion}}

On global gridded 2\,m temperature forecasts, per-cell conformal methods deliver near-nominal marginal coverage with no clear drift over three test years, pooled calibration is wasteful, and a Gaussian baseline is similarly sharp. Coverage on individual days is nonetheless far more variable across space than independence would imply, and misses cluster in time, with both effects growing with lead time and shared by all methods. The same holds for ECMWF HRES, a more skilful forecaster at all but the longest leads. An oracle day-scale accounts for most of the excess, but a day-scale predicted at issuance recovers only part of it and barely improves interval score, so on HRES no method is best on both day-level reliability and interval score from 12 hours to 3 days. These findings are limited to a reanalysis target, one variable and resolution, and two forecasters, one of them modest. They indicate that a dependence-aware calibration method, rather than further marginal adaptation, is the natural next step.

\section*{Data and code availability}

ERA5 is available through the WeatherBench 2 archive {[}15{]}. Code, regenerated result tables and figure scripts are available at \url{https://github.com/Mariyaaleem/conformal-wx}. The CNN forecasts used for the CNN results (\texttt{preds.npz}) are archived at \url{https://doi.org/10.5281/zenodo.23998807}; the ERA5 slice can be regenerated with the provided download script. Running \texttt{./run\_all.sh\ reproduce} regenerates the CNN tables and figures from \texttt{preds.npz}. The HRES forecasts (preds\_hres.npz) are ECMWF data and are not redistributed with the code; 08\_build\_hres\_preds.py rebuilds them from the WeatherBench 2 HRES archive on the same grid, dates and split labels, and the rebuilt values may differ slightly if that archive is updated. \texttt{./run\_all.sh\ hres} regenerates the HRES benchmark tables and the day-scale and paired analyses (decompose.py, feasible.py and paired\_hres.py; Sections 4.8 to 4.11 and Appendix G), and \texttt{./run\_all.sh\ dayscale} the day-scale analyses of the CNN forecasts.

\section*{Acknowledgements}

I thank my guide, Nadia Anjum, and Mohammed Izrar Uddin for their help. This work contains modified Copernicus Climate Change Service information 2026. Neither the European Commission nor ECMWF is responsible for any use that may be made of the Copernicus information or data it contains. ERA5 {[}13, 35, 36{]} was obtained through the WeatherBench 2 archive {[}15{]}, which provides a copy regridded to 5.625°. IFS HRES forecasts were obtained from the WeatherBench 2 archive {[}15{]}.

\section*{AI assistance}

The main results presented in this paper were obtained by the author. I used a generative AI assistant to improve the writing, review mathematical arguments, and check references. The author takes responsibility for the final content of this work.

\hypertarget{references}{%
\section*{References}\label{references}}

{[}1{]} V.~Vovk, A.~Gammerman, G.~Shafer. \emph{Algorithmic Learning in a Random World}. Springer, 2005.

{[}2{]} H.~Papadopoulos, K.~Proedrou, V.~Vovk, A.~Gammerman. Inductive confidence machines for regression. \emph{ECML}, 2002.

{[}3{]} J.~Lei, M.~G'Sell, A.~Rinaldo, R.~J.~Tibshirani, L.~Wasserman. Distribution-free predictive inference for regression. \emph{Journal of the American Statistical Association}, 113(523):1094--1111, 2018.

{[}4{]} A.~N.~Angelopoulos, S.~Bates. Conformal prediction: a gentle introduction. \emph{Foundations and Trends in Machine Learning}, 16(4):494--591, 2023.

{[}5{]} I.~Gibbs, E.~Candès. Adaptive conformal inference under distribution shift. \emph{NeurIPS}, 2021.

{[}6{]} M.~Zaffran, O.~Féron, Y.~Goude, J.~Josse, A.~Dieuleveut. Adaptive conformal predictions for time series. \emph{ICML}, PMLR 162:25834--25866, 2022.

{[}7{]} R.~F.~Barber, E.~J.~Candès, A.~Ramdas, R.~J.~Tibshirani. Conformal prediction beyond exchangeability. \emph{Annals of Statistics}, 51(2):816--845, 2023.

{[}8{]} C.~Xu, Y.~Xie. Conformal prediction interval for dynamic time-series. \emph{ICML}, PMLR 139:11559--11569, 2021.

{[}9{]} Z.~Lin, S.~Trivedi, J.~Sun. Conformal prediction intervals with temporal dependence. \emph{Transactions on Machine Learning Research}, 2022 (arXiv:2205.12940).

{[}10{]} R.~I.~Oliveira, P.~Orenstein, T.~Ramos, J.~V.~Romano. Split conformal prediction and non-exchangeable data. \emph{Journal of Machine Learning Research}, 25(225):1--38, 2024.

{[}11{]} H.~Mao, R.~Martin, B.~J.~Reich. Valid model-free spatial prediction. \emph{Journal of the American Statistical Association}, 119(546):904--914, 2024.

{[}12{]} S.~Messoudi, S.~Destercke, S.~Rousseau. Copula-based conformal prediction for multi-target regression. \emph{Pattern Recognition}, 2021, article 108101, doi:10.1016/j.patcog.2021.108101 (arXiv:2101.12002).

{[}13{]} H.~Hersbach et al. The ERA5 global reanalysis. \emph{Quarterly Journal of the Royal Meteorological Society}, 146(730):1999--2049, 2020.

{[}14{]} S.~Rasp, P.~D.~Dueben, S.~Scher, J.~A.~Weyn, S.~Mouatadid, N.~Thuerey. WeatherBench: a benchmark dataset for data-driven weather forecasting. \emph{Journal of Advances in Modeling Earth Systems}, 12(11), 2020.

{[}15{]} S.~Rasp et al. WeatherBench~2: a benchmark for the next generation of data-driven global weather models. \emph{Journal of Advances in Modeling Earth Systems}, 16(6), 2024, doi:10.1029/2023MS004019.

{[}16{]} S.~Garg, S.~Rasp, N.~Thuerey. WeatherBench Probability: a benchmark dataset for probabilistic medium-range weather forecasting along with deep learning baseline models. arXiv:2205.00865, 2022.

{[}17{]} V.~Gopakumar, A.~Gray, J.~Oskarsson, L. Zanisi, D. Giles, M. J. Kusner, S. Pamela, M. P. Deisenroth. Uncertainty quantification of surrogate models using conformal prediction. Machine Learning: Science and Technology, 7(1):015025, 2026, doi:10.1088/2632-2153/ae2e7b (arXiv:2408.09881).

{[}18{]} A.~Brusaferri, A.~Ballarino. Km-scale dynamical downscaling through conformalized latent diffusion models. arXiv:2510.13301, 2025.

{[}19{]} T.~Harris, R.~Sriver. Quantifying uncertainty in climate projections with conformal ensembles. arXiv:2408.06642, 2024.

{[}20{]} T.~Gneiting, A.~E.~Raftery. Strictly proper scoring rules, prediction, and estimation. \emph{Journal of the American Statistical Association}, 102(477):359--378, 2007.

{[}21{]} T.~Gneiting, F.~Balabdaoui, A.~E.~Raftery. Probabilistic forecasts, calibration and sharpness. \emph{Journal of the Royal Statistical Society B}, 69(2):243--268, 2007.

{[}22{]} T.~M.~Hamill. Interpretation of rank histograms for verifying ensemble forecasts. \emph{Monthly Weather Review}, 129(3):550--560, 2001.

{[}23{]} R.~E.~Livezey, W.~Y.~Chen. Statistical field significance and its determination by Monte Carlo techniques. \emph{Monthly Weather Review}, 111(1):46--59, 1983.

{[}24{]} D.~S.~Wilks. ``The stippling shows statistically significant grid points'': how research results are routinely overstated and overinterpreted, and what to do about it. \emph{Bulletin of the American Meteorological Society}, 97(12):2263--2273, 2016.

{[}25{]} A.~Asch, R.~Rossellini, P.~Hassanzadeh, R.~Willett. Rigorous uncertainty quantification of probabilistic AI weather forecasts with conformal prediction. arXiv:2606.19642, 2026.

{[}26{]} B.~R.~Feng, D.~K.~Park, X.~Luo, A.~Urdangarin, S.~Yoo, B.~J.~Reich. STACI: spatio-temporal aleatoric conformal inference. NeurIPS, 2025 (arXiv:2505.21658).

{[}27{]} T.~Gneiting, A.~E.~Raftery, A.~H.~Westveld, T.~Goldman. Calibrated probabilistic forecasting using ensemble model output statistics and minimum CRPS estimation. Monthly Weather Review, 133(5):1098--1118, 2005.

{[}28{]} M.~Clark, S.~Gangopadhyay, L.~Hay, B.~Rajagopalan, R.~Wilby. The Schaake shuffle: a method for reconstructing space--time variability in forecasted precipitation and temperature fields. Journal of Hydrometeorology, 5(1):243--262, 2004.

{[}29{]} R.~Schefzik, T.~L.~Thorarinsdottir, T.~Gneiting. Uncertainty quantification in complex simulation models using ensemble copula coupling. Statistical Science, 28(4):616--640, 2013.

{[}30{]} ECMWF. Implementation of IFS Cycle 46r1 (11 June 2019). \url{https://www.ecmwf.int/node/19151}

{[}31{]} ECMWF. Implementation of IFS cycle 45r1 (5 June 2018). \url{https://confluence.ecmwf.int/spaces/FCST/pages/97385120/Implementation+of+IFS+cycle+45r1}

{[}32{]} ECMWF. Changes to the ECMWF model, including cycle 47r3 (12 October 2021). \url{https://www.ecmwf.int/en/forecasts/documentation-and-support/changes-ecmwf-model}

{[}33{]} ECMWF. Implementation of IFS Cycle 47r1 (30 June 2020). \url{https://confluence.ecmwf.int/spaces/FCST/pages/179736603/Implementation+of+IFS+Cycle+47r1}

{[}34{]} M. Rodwell, M. Diamantakis, P. Düben, M. Janošek, S. Lang, I. Polichtchouk, F. Prates, C. Roberts, F. Váňa. IFS upgrade provides more skilful ensemble forecasts (IFS cycle 47r2, implemented 11 May 2021). ECMWF Newsletter 168, 18--23, 2021. doi:10.21957/m830hnz27r

{[}35{]} Copernicus Climate Change Service (2023). ERA5 hourly data on single levels from 1940 to present. Copernicus Climate Change Service (C3S) Climate Data Store (CDS). DOI: 10.24381/cds.adbb2d47.

{[}36{]} Copernicus Climate Change Service (2023). ERA5 hourly data on pressure levels from 1940 to present. Copernicus Climate Change Service (C3S) Climate Data Store (CDS). DOI: 10.24381/cds.bd0915c6.

\appendix
\hypertarget{appendix-a.-reproducibility}{%
\section{Reproducibility}\label{appendix-a.-reproducibility}}

The pipeline consists of: a synthetic-data generator used only for validating the diagnostics; a script that downloads the ERA5 slice from the WeatherBench 2 store; a script that trains the forecaster and writes calibration and test forecasts; a benchmark runner that builds intervals for all procedures and writes summary tables; a script that computes the block-bootstrap and paired tables; a script for the heterogeneity-aware overdispersion index, the threshold sweep and the bias-by-year table; a figure script; and a check script that compares regenerated tables with the reference tables. The repository also contains a script that builds the HRES forecasts from the WeatherBench 2 archive, the day-scale decomposition, the day-scale predictability test, the paired comparison of the day-scale model with the benchmark methods, and a script for the area-weighted test RMSE of both forecasters. All forecaster hyperparameters are in Section 3.1 and the following subsection; the random seed was 0. Interval-score and coverage tables with bootstrap intervals were computed from per-day area-weighted coverage and per-day interval scores stored by the benchmark runner. Code is at \url{https://github.com/Mariyaaleem/conformal-wx}; the CNN forecasts (preds.npz) are archived at \url{https://doi.org/10.5281/zenodo.23998807}. The HRES forecasts are ECMWF data and are not redistributed; 08\_build\_hres\_preds.py rebuilds them from the WeatherBench 2 archive. Given the CNN forecasts, ./run\_all.sh reproduce regenerates the tables and figures of the CNN benchmark; given the rebuilt HRES forecasts, ./run\_all.sh hres regenerates the HRES tables and the day-scale and paired analyses, and ./run\_all.sh dayscale those of the CNN. Retraining the forecaster on a GPU is not bit-reproducible, so the archived CNN forecasts are the canonical input for the CNN results.

\hypertarget{appendix-b.-homogeneous-versus-heterogeneity-aware-overdispersion}{%
\section{Homogeneous versus heterogeneity-aware overdispersion}\label{appendix-b.-homogeneous-versus-heterogeneity-aware-overdispersion}}

\Needspace{9\baselineskip}\medskip\noindent\textbf{Table 6.} $\mathrm{VIF}^{\mathrm{hom}}$ (common miss rate) and VIF (per-cell miss rates) at α\,=\,0.1.

\nopagebreak
\begin{center}
\small
\begin{tabular}{lcccccc}
\toprule
 & \multicolumn{3}{c}{$\mathrm{VIF}^{\mathrm{hom}}$} & \multicolumn{3}{c}{VIF (heterogeneity-aware)} \\
\cmidrule(lr){2-4}\cmidrule(lr){5-7}
Method & 24\,h & 120\,h & 240\,h & 24\,h & 120\,h & 240\,h \\
\midrule
Gaussian (RMSE) & 3.85 & 6.09 & 6.51 & 3.86 & 6.10 & 6.53 \\
Split, pooled & 5.74 & 7.17 & 7.62 & 9.27 & 12.66 & 13.79 \\
Split, per-cell & 3.84 & 6.04 & 6.48 & 3.85 & 6.03 & 6.47 \\
Per-cell + season & 3.31 & 5.10 & 5.30 & 3.31 & 5.09 & 5.30 \\
ACI (γ\,=\,0.005) & 3.59 & 6.65 & 7.60 & 3.59 & 6.65 & 7.59 \\
ACI (γ\,=\,0.02) & 3.74 & 6.98 & 8.44 & 3.74 & 6.97 & 8.43 \\
Rolling 90\,d & 4.37 & 7.88 & 10.18 & 4.40 & 7.98 & 10.33 \\
Rolling 365\,d & 3.85 & 6.60 & 7.01 & 3.85 & 6.60 & 7.00 \\
\bottomrule
\end{tabular}
\end{center}

For pooled split conformal prediction, which has very unequal per-cell miss rates, the heterogeneity-aware version is larger; for the other methods the two agree within 0.15. Bootstrap intervals in Table~2 and Table~5 are for \(\mathrm{VIF}^{\mathrm{hom}}\); I did not compute bootstrap intervals for the heterogeneity-aware version.

\hypertarget{appendix-c.-forecast-bias-by-year}{%
\section{Forecast bias by year}\label{appendix-c.-forecast-bias-by-year}}

\Needspace{9\baselineskip}\medskip\noindent\textbf{Table 7.} Mean forecast error y\,−\,\ensuremath{\hat{y}} (K), unweighted over cells, by year of initialisation. 2017–2018 are calibration years; 2019–2021 are test years.

\nopagebreak
\begin{longtable}[]{@{}
  >{\raggedright\arraybackslash}p{(\columnwidth - 16\tabcolsep) * \real{0.1111}}
  >{\raggedright\arraybackslash}p{(\columnwidth - 16\tabcolsep) * \real{0.1111}}
  >{\raggedright\arraybackslash}p{(\columnwidth - 16\tabcolsep) * \real{0.1111}}
  >{\raggedright\arraybackslash}p{(\columnwidth - 16\tabcolsep) * \real{0.1111}}
  >{\raggedright\arraybackslash}p{(\columnwidth - 16\tabcolsep) * \real{0.1111}}
  >{\raggedright\arraybackslash}p{(\columnwidth - 16\tabcolsep) * \real{0.1111}}
  >{\raggedright\arraybackslash}p{(\columnwidth - 16\tabcolsep) * \real{0.1111}}
  >{\raggedright\arraybackslash}p{(\columnwidth - 16\tabcolsep) * \real{0.1111}}
  >{\raggedright\arraybackslash}p{(\columnwidth - 16\tabcolsep) * \real{0.1111}}@{}}
\toprule\noalign{}
\begin{minipage}[b]{\linewidth}\raggedright
Year
\end{minipage} & \begin{minipage}[b]{\linewidth}\raggedright
6\,h
\end{minipage} & \begin{minipage}[b]{\linewidth}\raggedright
12\,h
\end{minipage} & \begin{minipage}[b]{\linewidth}\raggedright
1\,d
\end{minipage} & \begin{minipage}[b]{\linewidth}\raggedright
2\,d
\end{minipage} & \begin{minipage}[b]{\linewidth}\raggedright
3\,d
\end{minipage} & \begin{minipage}[b]{\linewidth}\raggedright
5\,d
\end{minipage} & \begin{minipage}[b]{\linewidth}\raggedright
7\,d
\end{minipage} & \begin{minipage}[b]{\linewidth}\raggedright
10\,d
\end{minipage} \\
\midrule\noalign{}
\endfirsthead
\toprule\noalign{}
\begin{minipage}[b]{\linewidth}\raggedright
Year
\end{minipage} & \begin{minipage}[b]{\linewidth}\raggedright
6\,h
\end{minipage} & \begin{minipage}[b]{\linewidth}\raggedright
12\,h
\end{minipage} & \begin{minipage}[b]{\linewidth}\raggedright
1\,d
\end{minipage} & \begin{minipage}[b]{\linewidth}\raggedright
2\,d
\end{minipage} & \begin{minipage}[b]{\linewidth}\raggedright
3\,d
\end{minipage} & \begin{minipage}[b]{\linewidth}\raggedright
5\,d
\end{minipage} & \begin{minipage}[b]{\linewidth}\raggedright
7\,d
\end{minipage} & \begin{minipage}[b]{\linewidth}\raggedright
10\,d
\end{minipage} \\
\midrule\noalign{}
\endhead
\bottomrule\noalign{}
\endlastfoot
2017 & 0.390 & 0.269 & \,−\,0.074 & \,−\,0.099 & \,−\,0.155 & \,−\,0.086 & 0.021 & 0.151 \\
2018 & 0.417 & 0.300 & \,−\,0.062 & \,−\,0.080 & \,−\,0.128 & \,−\,0.044 & 0.069 & 0.197 \\
2019 & 0.413 & 0.295 & \,−\,0.058 & \,−\,0.080 & \,−\,0.139 & \,−\,0.065 & 0.050 & 0.193 \\
2020 & 0.401 & 0.303 & \,−\,0.064 & \,−\,0.083 & \,−\,0.138 & \,−\,0.064 & 0.047 & 0.179 \\
2021 & 0.403 & 0.282 & \,−\,0.087 & \,−\,0.133 & \,−\,0.196 & \,−\,0.146 & \,−\,0.058 & 0.047 \\
\end{longtable}

\hypertarget{appendix-d.-share-of-forecasts-with-low-daily-coverage}{%
\section{Share of forecasts with low daily coverage}\label{appendix-d.-share-of-forecasts-with-low-daily-coverage}}

\Needspace{9\baselineskip}\medskip\noindent\textbf{Table 8.} Per-cell split conformal prediction, α\,=\,0.1: share of test forecasts whose area-weighted daily coverage is below 0.90\,−\,δ.

\nopagebreak
\begin{longtable}[]{@{}
  >{\raggedright\arraybackslash}p{(\columnwidth - 16\tabcolsep) * \real{0.1111}}
  >{\raggedright\arraybackslash}p{(\columnwidth - 16\tabcolsep) * \real{0.1111}}
  >{\raggedright\arraybackslash}p{(\columnwidth - 16\tabcolsep) * \real{0.1111}}
  >{\raggedright\arraybackslash}p{(\columnwidth - 16\tabcolsep) * \real{0.1111}}
  >{\raggedright\arraybackslash}p{(\columnwidth - 16\tabcolsep) * \real{0.1111}}
  >{\raggedright\arraybackslash}p{(\columnwidth - 16\tabcolsep) * \real{0.1111}}
  >{\raggedright\arraybackslash}p{(\columnwidth - 16\tabcolsep) * \real{0.1111}}
  >{\raggedright\arraybackslash}p{(\columnwidth - 16\tabcolsep) * \real{0.1111}}
  >{\raggedright\arraybackslash}p{(\columnwidth - 16\tabcolsep) * \real{0.1111}}@{}}
\toprule\noalign{}
\begin{minipage}[b]{\linewidth}\raggedright
δ
\end{minipage} & \begin{minipage}[b]{\linewidth}\raggedright
6\,h
\end{minipage} & \begin{minipage}[b]{\linewidth}\raggedright
12\,h
\end{minipage} & \begin{minipage}[b]{\linewidth}\raggedright
1\,d
\end{minipage} & \begin{minipage}[b]{\linewidth}\raggedright
2\,d
\end{minipage} & \begin{minipage}[b]{\linewidth}\raggedright
3\,d
\end{minipage} & \begin{minipage}[b]{\linewidth}\raggedright
5\,d
\end{minipage} & \begin{minipage}[b]{\linewidth}\raggedright
7\,d
\end{minipage} & \begin{minipage}[b]{\linewidth}\raggedright
10\,d
\end{minipage} \\
\midrule\noalign{}
\endfirsthead
\toprule\noalign{}
\begin{minipage}[b]{\linewidth}\raggedright
δ
\end{minipage} & \begin{minipage}[b]{\linewidth}\raggedright
6\,h
\end{minipage} & \begin{minipage}[b]{\linewidth}\raggedright
12\,h
\end{minipage} & \begin{minipage}[b]{\linewidth}\raggedright
1\,d
\end{minipage} & \begin{minipage}[b]{\linewidth}\raggedright
2\,d
\end{minipage} & \begin{minipage}[b]{\linewidth}\raggedright
3\,d
\end{minipage} & \begin{minipage}[b]{\linewidth}\raggedright
5\,d
\end{minipage} & \begin{minipage}[b]{\linewidth}\raggedright
7\,d
\end{minipage} & \begin{minipage}[b]{\linewidth}\raggedright
10\,d
\end{minipage} \\
\midrule\noalign{}
\endhead
\bottomrule\noalign{}
\endlastfoot
0.02 & 0.097 & 0.147 & 0.134 & 0.179 & 0.207 & 0.228 & 0.250 & 0.273 \\
0.05 & 0.001 & 0.001 & 0.001 & 0.003 & 0.008 & 0.020 & 0.019 & 0.008 \\
0.10 & 0 & 0 & 0 & 0 & 0 & 0 & 0 & 0 \\
\end{longtable}

\hypertarget{appendix-e.-results-at-other-significance-levels}{%
\section{Results at other significance levels}\label{appendix-e.-results-at-other-significance-levels}}

Tables~9 and~10 give the 120\,h results for α\,=\,0.2 and 0.05 with the same definitions as Table~2; ``π\,−\,π\ensuremath{^{0}}'' is the persistence excess and ``5th pct'' the 5th percentile of daily spatial coverage.

\Needspace{9\baselineskip}\medskip\noindent\textbf{Table 9.} α\,=\,0.2 (nominal 0.80), lead 120\,h.

\nopagebreak
\begin{longtable}[]{@{}
  >{\raggedright\arraybackslash}p{(\columnwidth - 10\tabcolsep) * \real{0.1667}}
  >{\raggedright\arraybackslash}p{(\columnwidth - 10\tabcolsep) * \real{0.1667}}
  >{\raggedright\arraybackslash}p{(\columnwidth - 10\tabcolsep) * \real{0.1667}}
  >{\raggedright\arraybackslash}p{(\columnwidth - 10\tabcolsep) * \real{0.1667}}
  >{\raggedright\arraybackslash}p{(\columnwidth - 10\tabcolsep) * \real{0.1667}}
  >{\raggedright\arraybackslash}p{(\columnwidth - 10\tabcolsep) * \real{0.1667}}@{}}
\toprule\noalign{}
\begin{minipage}[b]{\linewidth}\raggedright
Method
\end{minipage} & \begin{minipage}[b]{\linewidth}\raggedright
Coverage
\end{minipage} & \begin{minipage}[b]{\linewidth}\raggedright
Interval score (K)
\end{minipage} & \begin{minipage}[b]{\linewidth}\raggedright
\(\mathrm{VIF}^{\mathrm{hom}}\)
\end{minipage} & \begin{minipage}[b]{\linewidth}\raggedright
π\,−\,π\ensuremath{^{0}}
\end{minipage} & \begin{minipage}[b]{\linewidth}\raggedright
5th pct
\end{minipage} \\
\midrule\noalign{}
\endfirsthead
\toprule\noalign{}
\begin{minipage}[b]{\linewidth}\raggedright
Method
\end{minipage} & \begin{minipage}[b]{\linewidth}\raggedright
Coverage
\end{minipage} & \begin{minipage}[b]{\linewidth}\raggedright
Interval score (K)
\end{minipage} & \begin{minipage}[b]{\linewidth}\raggedright
\(\mathrm{VIF}^{\mathrm{hom}}\)
\end{minipage} & \begin{minipage}[b]{\linewidth}\raggedright
π\,−\,π\ensuremath{^{0}}
\end{minipage} & \begin{minipage}[b]{\linewidth}\raggedright
5th pct
\end{minipage} \\
\midrule\noalign{}
\endhead
\bottomrule\noalign{}
\endlastfoot
Gaussian (RMSE) & 0.801 {[}0.796, 0.807{]} & 6.99 {[}6.82, 7.14{]} & 6.53 & 0.263 & 0.759 \\
Split, pooled & 0.867 {[}0.861, 0.873{]} & 9.29 {[}9.04, 9.53{]} & 6.69 & 0.211 & 0.834 \\
Split, per-cell & 0.792 {[}0.786, 0.798{]} & 6.98 {[}6.80, 7.15{]} & 6.82 & 0.262 & 0.749 \\
Per-cell + season & 0.790 {[}0.787, 0.794{]} & 6.66 {[}6.48, 6.83{]} & 5.47 & 0.235 & 0.751 \\
ACI (γ\,=\,0.005) & 0.800 {[}0.794, 0.806{]} & 6.95 {[}6.77, 7.12{]} & 7.12 & 0.265 & 0.758 \\
ACI (γ\,=\,0.02) & 0.800 {[}0.795, 0.806{]} & 6.79 {[}6.60, 6.96{]} & 6.70 & 0.253 & 0.758 \\
Rolling 90\,d & 0.790 {[}0.783, 0.797{]} & 6.91 {[}6.71, 7.09{]} & 8.03 & 0.263 & 0.742 \\
Rolling 365\,d & 0.798 {[}0.792, 0.804{]} & 6.98 {[}6.81, 7.15{]} & 7.19 & 0.268 & 0.754 \\
\end{longtable}

\Needspace{9\baselineskip}\medskip\noindent\textbf{Table 10.} α\,=\,0.05 (nominal 0.95), lead 120\,h.

\nopagebreak
\begin{longtable}[]{@{}
  >{\raggedright\arraybackslash}p{(\columnwidth - 10\tabcolsep) * \real{0.1667}}
  >{\raggedright\arraybackslash}p{(\columnwidth - 10\tabcolsep) * \real{0.1667}}
  >{\raggedright\arraybackslash}p{(\columnwidth - 10\tabcolsep) * \real{0.1667}}
  >{\raggedright\arraybackslash}p{(\columnwidth - 10\tabcolsep) * \real{0.1667}}
  >{\raggedright\arraybackslash}p{(\columnwidth - 10\tabcolsep) * \real{0.1667}}
  >{\raggedright\arraybackslash}p{(\columnwidth - 10\tabcolsep) * \real{0.1667}}@{}}
\toprule\noalign{}
\begin{minipage}[b]{\linewidth}\raggedright
Method
\end{minipage} & \begin{minipage}[b]{\linewidth}\raggedright
Coverage
\end{minipage} & \begin{minipage}[b]{\linewidth}\raggedright
Interval score (K)
\end{minipage} & \begin{minipage}[b]{\linewidth}\raggedright
\(\mathrm{VIF}^{\mathrm{hom}}\)
\end{minipage} & \begin{minipage}[b]{\linewidth}\raggedright
π\,−\,π\ensuremath{^{0}}
\end{minipage} & \begin{minipage}[b]{\linewidth}\raggedright
5th pct
\end{minipage} \\
\midrule\noalign{}
\endfirsthead
\toprule\noalign{}
\begin{minipage}[b]{\linewidth}\raggedright
Method
\end{minipage} & \begin{minipage}[b]{\linewidth}\raggedright
Coverage
\end{minipage} & \begin{minipage}[b]{\linewidth}\raggedright
Interval score (K)
\end{minipage} & \begin{minipage}[b]{\linewidth}\raggedright
\(\mathrm{VIF}^{\mathrm{hom}}\)
\end{minipage} & \begin{minipage}[b]{\linewidth}\raggedright
π\,−\,π\ensuremath{^{0}}
\end{minipage} & \begin{minipage}[b]{\linewidth}\raggedright
5th pct
\end{minipage} \\
\midrule\noalign{}
\endhead
\bottomrule\noalign{}
\endlastfoot
Gaussian (RMSE) & 0.941 {[}0.938, 0.943{]} & 9.78 {[}9.52, 10.04{]} & 5.82 & 0.276 & 0.915 \\
Split, pooled & 0.971 {[}0.968, 0.974{]} & 16.28 {[}15.84, 16.68{]} & 7.27 & 0.310 & 0.950 \\
Split, per-cell & 0.946 {[}0.943, 0.948{]} & 9.76 {[}9.54, 9.98{]} & 5.50 & 0.265 & 0.921 \\
Per-cell + season & 0.943 {[}0.941, 0.945{]} & 9.09 {[}8.84, 9.31{]} & 4.74 & 0.253 & 0.921 \\
ACI (γ\,=\,0.005) & 0.950 {[}0.948, 0.953{]} & 9.72 {[}9.49, 9.93{]} & 5.94 & 0.267 & 0.926 \\
ACI (γ\,=\,0.02) & 0.950 {[}0.948, 0.953{]} & 9.70 {[}9.42, 9.97{]} & 6.36 & 0.273 & 0.926 \\
Rolling 90\,d & 0.942 {[}0.938, 0.946{]} & 9.71 {[}9.41, 9.97{]} & 7.12 & 0.290 & 0.914 \\
Rolling 365\,d & 0.948 {[}0.946, 0.951{]} & 9.78 {[}9.56, 10.02{]} & 5.93 & 0.271 & 0.924 \\
\end{longtable}

\Needspace{9\baselineskip}\medskip\noindent\textbf{Table 11.} Paired differences versus per-cell split conformal prediction at other significance levels (95\% paired block-bootstrap intervals).

\nopagebreak
\begin{center}
\small
\begin{tabular}{lccccc}
\toprule
 &  & \multicolumn{2}{c}{Per-cell + season} & \multicolumn{2}{c}{ACI (γ\,=\,0.005)} \\
\cmidrule(lr){3-4}\cmidrule(lr){5-6}
α & Lead & Δ interval score & $\Delta\,\mathrm{VIF}^{\mathrm{hom}}$ & Δ interval score & $\Delta\,\mathrm{VIF}^{\mathrm{hom}}$ \\
\midrule
0.2 & 24\,h & [−0.146,−0.091] & [−1.65,−0.24] & [−0.038,−0.007] & [−0.46,0.17] \\
 & 120\,h & [−0.403,−0.228] & [−2.55,−0.39] & [−0.078,0.010] & [−0.24,0.82] \\
 & 240\,h & [−0.419,−0.244] & [−3.58,−0.91] & [−0.047,0.044] & [−0.14,1.45] \\
0.05 & 24\,h & [−0.357,−0.214] & [−0.89,0.01] & [−0.084,−0.016] & [−0.60,0.18] \\
 & 120\,h & [−0.872,−0.488] & [−2.09,0.24] & [−0.126,0.038] & [−0.34,1.02] \\
 & 240\,h & [−0.895,−0.495] & [−1.45,0.42] & [−0.052,0.123] & [0.04,1.58] \\
\bottomrule
\end{tabular}
\end{center}

\hypertarget{appendix-f.-synthetic-validation-of-the-diagnostics}{%
\section{Synthetic validation of the diagnostics}\label{appendix-f.-synthetic-validation-of-the-diagnostics}}

On exchangeable, spatially independent synthetic errors with 365 calibration and 730 test days, the average coverage over leads was between 0.899 and 0.902 for the Gaussian, pooled, per-cell and ACI (γ\,=\,0.005) procedures. For the Gaussian and pooled procedures, VIF over the eight leads ranged from 0.88 to 1.06 and persistence matched its shuffled-days null within 0.003. These synthetic results validate the diagnostics only and are not experimental findings.

\hypertarget{appendix-g.-day-scale-decomposition-predictability-and-second-forecaster-results}{%
\section{Day-scale decomposition, predictability and second-forecaster results}\label{appendix-g.-day-scale-decomposition-predictability-and-second-forecaster-results}}

All results are at α = 0.1 with per-cell split conformal as the base method, 95\% moving-block bootstrap intervals with 30-day blocks and 200 replicates for Table~12 (1,000 for Tables~14, 18 and 19), and the area-weighted overdispersion index of Section~3.6. ``Excess removed'' is (VIF base − VIF oracle) / (VIF base − VIF floor) for the global oracle. The split-half drop is the base index on the evaluation half minus the split-half index.

\Needspace{9\baselineskip}\medskip\noindent\textbf{Table 12.} CNN decomposition: oracle day-scale, null floor and split-half variant.

\nopagebreak
\begin{longtable}[]{@{}
  >{\raggedright\arraybackslash}p{(\columnwidth - 10\tabcolsep) * \real{0.0843}}
  >{\raggedright\arraybackslash}p{(\columnwidth - 10\tabcolsep) * \real{0.2229}}
  >{\raggedright\arraybackslash}p{(\columnwidth - 10\tabcolsep) * \real{0.1988}}
  >{\raggedright\arraybackslash}p{(\columnwidth - 10\tabcolsep) * \real{0.1084}}
  >{\raggedright\arraybackslash}p{(\columnwidth - 10\tabcolsep) * \real{0.1325}}
  >{\raggedright\arraybackslash}p{(\columnwidth - 10\tabcolsep) * \real{0.2530}}@{}}
\toprule\noalign{}
\begin{minipage}[b]{\linewidth}\raggedright
Lead
\end{minipage} & \begin{minipage}[b]{\linewidth}\raggedright
Base VIF {[}95\% CI{]}
\end{minipage} & \begin{minipage}[b]{\linewidth}\raggedright
Oracle VIF {[}95\% CI{]}
\end{minipage} & \begin{minipage}[b]{\linewidth}\raggedright
Oracle floor
\end{minipage} & \begin{minipage}[b]{\linewidth}\raggedright
Excess removed
\end{minipage} & \begin{minipage}[b]{\linewidth}\raggedright
Split-half drop {[}95\% CI{]}
\end{minipage} \\
\midrule\noalign{}
\endfirsthead
\toprule\noalign{}
\begin{minipage}[b]{\linewidth}\raggedright
Lead
\end{minipage} & \begin{minipage}[b]{\linewidth}\raggedright
Base VIF {[}95\% CI{]}
\end{minipage} & \begin{minipage}[b]{\linewidth}\raggedright
Oracle VIF {[}95\% CI{]}
\end{minipage} & \begin{minipage}[b]{\linewidth}\raggedright
Oracle floor
\end{minipage} & \begin{minipage}[b]{\linewidth}\raggedright
Excess removed
\end{minipage} & \begin{minipage}[b]{\linewidth}\raggedright
Split-half drop {[}95\% CI{]}
\end{minipage} \\
\midrule\noalign{}
\endhead
\bottomrule\noalign{}
\endlastfoot
6\,h & 3.69 {[}3.23, 4.17{]} & 1.14 {[}1.02, 1.23{]} & 0.55 & 81\% & 0.22 {[}−0.02, 0.41{]} \\
12\,h & 4.39 {[}3.65, 4.97{]} & 1.03 {[}0.90, 1.16{]} & 0.61 & 89\% & 0.76 {[}0.45, 1.06{]} \\
1\,d & 3.84 {[}3.19, 4.33{]} & 0.77 {[}0.70, 0.84{]} & 0.49 & 92\% & 0.68 {[}0.32, 0.94{]} \\
2\,d & 4.57 {[}3.90, 5.14{]} & 0.86 {[}0.78, 0.95{]} & 0.49 & 91\% & 0.98 {[}0.67, 1.25{]} \\
3\,d & 5.23 {[}4.50, 5.86{]} & 0.94 {[}0.83, 1.03{]} & 0.48 & 90\% & 1.26 {[}0.94, 1.54{]} \\
5\,d & 6.02 {[}4.85, 7.38{]} & 1.23 {[}1.08, 1.40{]} & 0.52 & 87\% & 1.48 {[}0.92, 2.09{]} \\
7\,d & 6.23 {[}5.19, 7.20{]} & 1.23 {[}1.05, 1.39{]} & 0.51 & 87\% & 1.56 {[}1.17, 2.06{]} \\
10\,d & 6.47 {[}5.60, 7.29{]} & 1.31 {[}1.10, 1.51{]} & 0.56 & 87\% & 1.67 {[}1.30, 2.09{]} \\
\end{longtable}

\Needspace{9\baselineskip}\medskip\noindent\textbf{Table 13.} Day-scale model of Section~3.9 against per-cell split conformal, for both forecasters. R² is out-of-sample on the test days; IS ratio is the interval score of the day-scale model divided by that of per-cell split. VIF is shown for the base method and the primary model.

\nopagebreak
\begin{longtable}[]{@{}
  >{\raggedright\arraybackslash}p{(\columnwidth - 12\tabcolsep) * \real{0.0920}}
  >{\raggedright\arraybackslash}p{(\columnwidth - 12\tabcolsep) * \real{0.1322}}
  >{\raggedright\arraybackslash}p{(\columnwidth - 12\tabcolsep) * \real{0.1954}}
  >{\raggedright\arraybackslash}p{(\columnwidth - 12\tabcolsep) * \real{0.1322}}
  >{\raggedright\arraybackslash}p{(\columnwidth - 12\tabcolsep) * \real{0.1264}}
  >{\raggedright\arraybackslash}p{(\columnwidth - 12\tabcolsep) * \real{0.1954}}
  >{\raggedright\arraybackslash}p{(\columnwidth - 12\tabcolsep) * \real{0.1264}}@{}}
\toprule\noalign{}
\begin{minipage}[b]{\linewidth}\raggedright
Lead
\end{minipage} & \begin{minipage}[b]{\linewidth}\raggedright
CNN R²
\end{minipage} & \begin{minipage}[b]{\linewidth}\raggedright
CNN VIF, base to primary
\end{minipage} & \begin{minipage}[b]{\linewidth}\raggedright
CNN IS ratio
\end{minipage} & \begin{minipage}[b]{\linewidth}\raggedright
HRES R²
\end{minipage} & \begin{minipage}[b]{\linewidth}\raggedright
HRES VIF, base to primary
\end{minipage} & \begin{minipage}[b]{\linewidth}\raggedright
HRES IS ratio
\end{minipage} \\
\midrule\noalign{}
\endfirsthead
\toprule\noalign{}
\begin{minipage}[b]{\linewidth}\raggedright
Lead
\end{minipage} & \begin{minipage}[b]{\linewidth}\raggedright
CNN R²
\end{minipage} & \begin{minipage}[b]{\linewidth}\raggedright
CNN VIF, base to primary
\end{minipage} & \begin{minipage}[b]{\linewidth}\raggedright
CNN IS ratio
\end{minipage} & \begin{minipage}[b]{\linewidth}\raggedright
HRES R²
\end{minipage} & \begin{minipage}[b]{\linewidth}\raggedright
HRES VIF, base to primary
\end{minipage} & \begin{minipage}[b]{\linewidth}\raggedright
HRES IS ratio
\end{minipage} \\
\midrule\noalign{}
\endhead
\bottomrule\noalign{}
\endlastfoot
6\,h & 0.31 & 3.69 to 3.01 & 0.999 & 0.67 & 9.01 to 3.37 & 0.996 \\
12\,h & 0.45 & 4.39 to 2.46 & 0.998 & 0.74 & 8.05 to 2.71 & 0.998 \\
1\,d & 0.51 & 3.84 to 2.46 & 0.997 & 0.62 & 3.89 to 1.95 & 0.997 \\
2\,d & 0.46 & 4.57 to 3.19 & 0.996 & 0.38 & 3.25 to 2.16 & 0.998 \\
3\,d & 0.40 & 5.23 to 3.71 & 0.996 & 0.17 & 3.08 to 2.40 & 0.999 \\
5\,d & 0.37 & 6.02 to 4.65 & 0.995 & 0.06 & 4.16 to 3.64 & 0.999 \\
7\,d & 0.31 & 6.23 to 4.70 & 0.995 & 0.12 & 5.93 to 5.31 & 0.999 \\
10\,d & 0.36 & 6.47 to 4.67 & 0.994 & 0.14 & 8.03 to 7.45 & 0.999 \\
\end{longtable}

\Needspace{9\baselineskip}\medskip\noindent\textbf{Table 14.} HRES decomposition: oracle day-scale, null floor and split-half variant.

\nopagebreak
\begin{longtable}[]{@{}
  >{\raggedright\arraybackslash}p{(\columnwidth - 10\tabcolsep) * \real{0.0843}}
  >{\raggedright\arraybackslash}p{(\columnwidth - 10\tabcolsep) * \real{0.2229}}
  >{\raggedright\arraybackslash}p{(\columnwidth - 10\tabcolsep) * \real{0.1988}}
  >{\raggedright\arraybackslash}p{(\columnwidth - 10\tabcolsep) * \real{0.1084}}
  >{\raggedright\arraybackslash}p{(\columnwidth - 10\tabcolsep) * \real{0.1325}}
  >{\raggedright\arraybackslash}p{(\columnwidth - 10\tabcolsep) * \real{0.2530}}@{}}
\toprule\noalign{}
\begin{minipage}[b]{\linewidth}\raggedright
Lead
\end{minipage} & \begin{minipage}[b]{\linewidth}\raggedright
Base VIF {[}95\% CI{]}
\end{minipage} & \begin{minipage}[b]{\linewidth}\raggedright
Oracle VIF {[}95\% CI{]}
\end{minipage} & \begin{minipage}[b]{\linewidth}\raggedright
Oracle floor
\end{minipage} & \begin{minipage}[b]{\linewidth}\raggedright
Excess removed
\end{minipage} & \begin{minipage}[b]{\linewidth}\raggedright
Split-half drop {[}95\% CI{]}
\end{minipage} \\
\midrule\noalign{}
\endfirsthead
\toprule\noalign{}
\begin{minipage}[b]{\linewidth}\raggedright
Lead
\end{minipage} & \begin{minipage}[b]{\linewidth}\raggedright
Base VIF {[}95\% CI{]}
\end{minipage} & \begin{minipage}[b]{\linewidth}\raggedright
Oracle VIF {[}95\% CI{]}
\end{minipage} & \begin{minipage}[b]{\linewidth}\raggedright
Oracle floor
\end{minipage} & \begin{minipage}[b]{\linewidth}\raggedright
Excess removed
\end{minipage} & \begin{minipage}[b]{\linewidth}\raggedright
Split-half drop {[}95\% CI{]}
\end{minipage} \\
\midrule\noalign{}
\endhead
\bottomrule\noalign{}
\endlastfoot
6\,h & 9.01 {[}2.70, 14.07{]} & 0.63 {[}0.56, 0.68{]} & 0.50 & 98\% & 3.26 {[}0.11, 5.77{]} \\
12\,h & 8.05 {[}3.64, 11.67{]} & 0.74 {[}0.68, 0.81{]} & 0.54 & 97\% & 2.82 {[}0.58, 4.66{]} \\
1\,d & 3.89 {[}3.00, 4.56{]} & 0.65 {[}0.60, 0.71{]} & 0.52 & 96\% & 0.68 {[}0.25, 1.05{]} \\
2\,d & 3.25 {[}2.81, 3.56{]} & 0.64 {[}0.58, 0.69{]} & 0.50 & 95\% & 0.31 {[}0.08, 0.49{]} \\
3\,d & 3.08 {[}2.72, 3.34{]} & 0.68 {[}0.61, 0.75{]} & 0.50 & 93\% & 0.17 {[}−0.07, 0.34{]} \\
5\,d & 4.16 {[}3.67, 4.56{]} & 0.64 {[}0.57, 0.70{]} & 0.51 & 96\% & 0.81 {[}0.58, 1.07{]} \\
7\,d & 5.93 {[}5.17, 6.64{]} & 0.68 {[}0.60, 0.72{]} & 0.45 & 96\% & 1.64 {[}1.26, 2.02{]} \\
10\,d & 8.03 {[}7.02, 8.88{]} & 0.83 {[}0.75, 0.90{]} & 0.43 & 95\% & 2.67 {[}2.22, 3.07{]} \\
\end{longtable}

\Needspace{9\baselineskip}\medskip\noindent\textbf{Table 15.} HRES overdispersion index (heterogeneity-aware) for the benchmark methods and the day-scale model; lower is better. Pooled split and the Gaussian baseline are omitted (pooled split over-covers).

\nopagebreak
\begin{longtable}[]{@{}
  >{\raggedright\arraybackslash}p{(\columnwidth - 12\tabcolsep) * \real{0.0936}}
  >{\raggedright\arraybackslash}p{(\columnwidth - 12\tabcolsep) * \real{0.1462}}
  >{\raggedright\arraybackslash}p{(\columnwidth - 12\tabcolsep) * \real{0.1462}}
  >{\raggedright\arraybackslash}p{(\columnwidth - 12\tabcolsep) * \real{0.1462}}
  >{\raggedright\arraybackslash}p{(\columnwidth - 12\tabcolsep) * \real{0.1462}}
  >{\raggedright\arraybackslash}p{(\columnwidth - 12\tabcolsep) * \real{0.1462}}
  >{\raggedright\arraybackslash}p{(\columnwidth - 12\tabcolsep) * \real{0.1754}}@{}}
\toprule\noalign{}
\begin{minipage}[b]{\linewidth}\raggedright
Lead
\end{minipage} & \begin{minipage}[b]{\linewidth}\raggedright
Split (per-cell)
\end{minipage} & \begin{minipage}[b]{\linewidth}\raggedright
Mondrian season
\end{minipage} & \begin{minipage}[b]{\linewidth}\raggedright
ACI (0.02)
\end{minipage} & \begin{minipage}[b]{\linewidth}\raggedright
Rolling 90
\end{minipage} & \begin{minipage}[b]{\linewidth}\raggedright
Rolling 365
\end{minipage} & \begin{minipage}[b]{\linewidth}\raggedright
Day-scale model
\end{minipage} \\
\midrule\noalign{}
\endfirsthead
\toprule\noalign{}
\begin{minipage}[b]{\linewidth}\raggedright
Lead
\end{minipage} & \begin{minipage}[b]{\linewidth}\raggedright
Split (per-cell)
\end{minipage} & \begin{minipage}[b]{\linewidth}\raggedright
Mondrian season
\end{minipage} & \begin{minipage}[b]{\linewidth}\raggedright
ACI (0.02)
\end{minipage} & \begin{minipage}[b]{\linewidth}\raggedright
Rolling 90
\end{minipage} & \begin{minipage}[b]{\linewidth}\raggedright
Rolling 365
\end{minipage} & \begin{minipage}[b]{\linewidth}\raggedright
Day-scale model
\end{minipage} \\
\midrule\noalign{}
\endhead
\bottomrule\noalign{}
\endlastfoot
6\,h & 9.02 & 9.23 & 3.91 & 4.54 & 6.87 & 3.37 \\
12\,h & 8.05 & 8.49 & 4.91 & 5.64 & 7.11 & 2.71 \\
1\,d & 3.89 & 3.57 & 3.38 & 3.41 & 3.75 & 1.95 \\
2\,d & 3.25 & 2.86 & 3.09 & 3.24 & 3.15 & 2.16 \\
3\,d & 3.08 & 2.86 & 3.10 & 3.10 & 3.02 & 2.40 \\
5\,d & 4.16 & 3.81 & 3.92 & 4.02 & 4.01 & 3.64 \\
7\,d & 5.94 & 5.54 & 5.66 & 5.98 & 5.75 & 5.31 \\
10\,d & 8.04 & 7.38 & 8.11 & 8.63 & 7.77 & 7.45 \\
\end{longtable}

\Needspace{9\baselineskip}\medskip\noindent\textbf{Table 16.} HRES mean interval score (K), area-weighted; lower is better.

\nopagebreak
\begin{longtable}[]{@{}
  >{\raggedright\arraybackslash}p{(\columnwidth - 12\tabcolsep) * \real{0.0936}}
  >{\raggedright\arraybackslash}p{(\columnwidth - 12\tabcolsep) * \real{0.1462}}
  >{\raggedright\arraybackslash}p{(\columnwidth - 12\tabcolsep) * \real{0.1462}}
  >{\raggedright\arraybackslash}p{(\columnwidth - 12\tabcolsep) * \real{0.1462}}
  >{\raggedright\arraybackslash}p{(\columnwidth - 12\tabcolsep) * \real{0.1462}}
  >{\raggedright\arraybackslash}p{(\columnwidth - 12\tabcolsep) * \real{0.1462}}
  >{\raggedright\arraybackslash}p{(\columnwidth - 12\tabcolsep) * \real{0.1754}}@{}}
\toprule\noalign{}
\begin{minipage}[b]{\linewidth}\raggedright
Lead
\end{minipage} & \begin{minipage}[b]{\linewidth}\raggedright
Split (per-cell)
\end{minipage} & \begin{minipage}[b]{\linewidth}\raggedright
Mondrian season
\end{minipage} & \begin{minipage}[b]{\linewidth}\raggedright
ACI (0.02)
\end{minipage} & \begin{minipage}[b]{\linewidth}\raggedright
Rolling 90
\end{minipage} & \begin{minipage}[b]{\linewidth}\raggedright
Rolling 365
\end{minipage} & \begin{minipage}[b]{\linewidth}\raggedright
Day-scale model
\end{minipage} \\
\midrule\noalign{}
\endfirsthead
\toprule\noalign{}
\begin{minipage}[b]{\linewidth}\raggedright
Lead
\end{minipage} & \begin{minipage}[b]{\linewidth}\raggedright
Split (per-cell)
\end{minipage} & \begin{minipage}[b]{\linewidth}\raggedright
Mondrian season
\end{minipage} & \begin{minipage}[b]{\linewidth}\raggedright
ACI (0.02)
\end{minipage} & \begin{minipage}[b]{\linewidth}\raggedright
Rolling 90
\end{minipage} & \begin{minipage}[b]{\linewidth}\raggedright
Rolling 365
\end{minipage} & \begin{minipage}[b]{\linewidth}\raggedright
Day-scale model
\end{minipage} \\
\midrule\noalign{}
\endhead
\bottomrule\noalign{}
\endlastfoot
6\,h & 1.866 & 1.805 & 1.733 & 1.774 & 1.814 & 1.859 \\
12\,h & 2.032 & 1.955 & 1.891 & 1.948 & 1.989 & 2.028 \\
1\,d & 2.142 & 2.047 & 1.993 & 2.045 & 2.101 & 2.136 \\
2\,d & 2.506 & 2.401 & 2.389 & 2.428 & 2.474 & 2.500 \\
3\,d & 2.959 & 2.840 & 2.864 & 2.896 & 2.935 & 2.955 \\
5\,d & 4.329 & 4.162 & 4.270 & 4.290 & 4.315 & 4.325 \\
7\,d & 6.099 & 5.838 & 6.060 & 6.076 & 6.090 & 6.092 \\
10\,d & 8.315 & 7.891 & 8.304 & 8.333 & 8.323 & 8.304 \\
\end{longtable}

Table~15 and 16 values for the benchmark methods come from the repository tables; the day-scale values come from my own evaluation code, whose base interval scores equal the repository's per-cell split scores at all eight leads. Tables~18 and 19 give paired comparisons with the day-scale model, using the repository's own interval construction. The recomputed values matched the saved results (interval scores exactly; VIF within 0.009, the difference expected from sample against population variance).

\Needspace{9\baselineskip}\medskip\noindent\textbf{Table 17.} HRES per-cell split coverage by test year, and the base index after removing each year’s mean daily coverage.

\nopagebreak
\begin{longtable}[]{@{}
  >{\raggedright\arraybackslash}p{(\columnwidth - 10\tabcolsep) * \real{0.0909}}
  >{\raggedright\arraybackslash}p{(\columnwidth - 10\tabcolsep) * \real{0.1364}}
  >{\raggedright\arraybackslash}p{(\columnwidth - 10\tabcolsep) * \real{0.1364}}
  >{\raggedright\arraybackslash}p{(\columnwidth - 10\tabcolsep) * \real{0.1364}}
  >{\raggedright\arraybackslash}p{(\columnwidth - 10\tabcolsep) * \real{0.2273}}
  >{\raggedright\arraybackslash}p{(\columnwidth - 10\tabcolsep) * \real{0.2727}}@{}}
\toprule\noalign{}
\begin{minipage}[b]{\linewidth}\raggedright
Lead
\end{minipage} & \begin{minipage}[b]{\linewidth}\raggedright
2019
\end{minipage} & \begin{minipage}[b]{\linewidth}\raggedright
2020
\end{minipage} & \begin{minipage}[b]{\linewidth}\raggedright
2021
\end{minipage} & \begin{minipage}[b]{\linewidth}\raggedright
Base VIF
\end{minipage} & \begin{minipage}[b]{\linewidth}\raggedright
VIF after removing year means
\end{minipage} \\
\midrule\noalign{}
\endfirsthead
\toprule\noalign{}
\begin{minipage}[b]{\linewidth}\raggedright
Lead
\end{minipage} & \begin{minipage}[b]{\linewidth}\raggedright
2019
\end{minipage} & \begin{minipage}[b]{\linewidth}\raggedright
2020
\end{minipage} & \begin{minipage}[b]{\linewidth}\raggedright
2021
\end{minipage} & \begin{minipage}[b]{\linewidth}\raggedright
Base VIF
\end{minipage} & \begin{minipage}[b]{\linewidth}\raggedright
VIF after removing year means
\end{minipage} \\
\midrule\noalign{}
\endhead
\bottomrule\noalign{}
\endlastfoot
6\,h & 0.886 & 0.879 & 0.864 & 9.01 & 7.63 \\
12\,h & 0.890 & 0.892 & 0.881 & 8.05 & 7.67 \\
1\,d & 0.890 & 0.894 & 0.885 & 3.89 & 3.65 \\
2\,d & 0.895 & 0.900 & 0.894 & 3.25 & 3.13 \\
3\,d & 0.899 & 0.905 & 0.899 & 3.08 & 2.91 \\
5\,d & 0.903 & 0.909 & 0.903 & 4.16 & 3.99 \\
7\,d & 0.905 & 0.910 & 0.903 & 5.93 & 5.79 \\
10\,d & 0.904 & 0.905 & 0.900 & 8.03 & 7.94 \\
\end{longtable}

The CNN's per-cell split coverage is flat across the three test years at every lead (0.892 to 0.900).

\Needspace{9\baselineskip}\medskip\noindent\textbf{Table 18.} HRES, alpha = 0.1: VIF of each benchmark method minus VIF of the day-scale model, with 95\% paired block-bootstrap intervals (1,000 replicates, 30-day blocks). Positive means the day-scale model has the lower index; * marks intervals that exclude zero. Methods differ in marginal coverage (0.874 to 0.907), so read the differences alongside coverage.

\nopagebreak
\begin{longtable}[]{@{}
  >{\raggedright\arraybackslash}p{(\columnwidth - 12\tabcolsep) * \real{0.0780}}
  >{\raggedright\arraybackslash}p{(\columnwidth - 12\tabcolsep) * \real{0.1537}}
  >{\raggedright\arraybackslash}p{(\columnwidth - 12\tabcolsep) * \real{0.1537}}
  >{\raggedright\arraybackslash}p{(\columnwidth - 12\tabcolsep) * \real{0.1537}}
  >{\raggedright\arraybackslash}p{(\columnwidth - 12\tabcolsep) * \real{0.1537}}
  >{\raggedright\arraybackslash}p{(\columnwidth - 12\tabcolsep) * \real{0.1537}}
  >{\raggedright\arraybackslash}p{(\columnwidth - 12\tabcolsep) * \real{0.1537}}@{}}
\toprule\noalign{}
\begin{minipage}[b]{\linewidth}\raggedright
Lead
\end{minipage} & \begin{minipage}[b]{\linewidth}\raggedright
Split (per-cell)
\end{minipage} & \begin{minipage}[b]{\linewidth}\raggedright
Mondrian season
\end{minipage} & \begin{minipage}[b]{\linewidth}\raggedright
ACI (0.005)
\end{minipage} & \begin{minipage}[b]{\linewidth}\raggedright
ACI (0.02)
\end{minipage} & \begin{minipage}[b]{\linewidth}\raggedright
Rolling 90
\end{minipage} & \begin{minipage}[b]{\linewidth}\raggedright
Rolling 365
\end{minipage} \\
\midrule\noalign{}
\endfirsthead
\toprule\noalign{}
\begin{minipage}[b]{\linewidth}\raggedright
Lead
\end{minipage} & \begin{minipage}[b]{\linewidth}\raggedright
Split (per-cell)
\end{minipage} & \begin{minipage}[b]{\linewidth}\raggedright
Mondrian season
\end{minipage} & \begin{minipage}[b]{\linewidth}\raggedright
ACI (0.005)
\end{minipage} & \begin{minipage}[b]{\linewidth}\raggedright
ACI (0.02)
\end{minipage} & \begin{minipage}[b]{\linewidth}\raggedright
Rolling 90
\end{minipage} & \begin{minipage}[b]{\linewidth}\raggedright
Rolling 365
\end{minipage} \\
\midrule\noalign{}
\endhead
\bottomrule\noalign{}
\endlastfoot
6\,h & +5.64 {[}+0.61, +9.77{]}* & +5.85 {[}+0.72, +8.85{]}* & +2.66 {[}+0.38, +5.59{]}* & +0.54 {[}−0.51, +2.38{]} & +1.16 {[}−0.12, +3.68{]} & +3.49 {[}+0.34, +6.82{]}* \\
12\,h & +5.34 {[}+1.33, +8.48{]}* & +5.78 {[}+1.81, +8.70{]}* & +4.55 {[}+1.05, +7.95{]}* & +2.20 {[}+0.56, +4.80{]}* & +2.93 {[}+0.71, +6.10{]}* & +4.40 {[}+0.85, +7.53{]}* \\
1\,d & +1.94 {[}+1.14, +2.46{]}* & +1.62 {[}+0.70, +2.39{]}* & +2.21 {[}+1.17, +2.98{]}* & +1.43 {[}+0.55, +2.26{]}* & +1.46 {[}+0.58, +2.34{]}* & +1.80 {[}+1.06, +2.34{]}* \\
2\,d & +1.09 {[}+0.69, +1.35{]}* & +0.70 {[}+0.29, +1.10{]}* & +1.27 {[}+0.66, +1.64{]}* & +0.92 {[}+0.33, +1.44{]}* & +1.08 {[}+0.43, +1.68{]}* & +0.99 {[}+0.60, +1.22{]}* \\
3\,d & +0.68 {[}+0.37, +0.91{]}* & +0.46 {[}+0.17, +0.83{]}* & +0.80 {[}+0.31, +1.11{]}* & +0.69 {[}+0.25, +1.06{]}* & +0.69 {[}+0.24, +1.10{]}* & +0.61 {[}+0.27, +0.91{]}* \\
5\,d & +0.51 {[}+0.16, +0.90{]}* & +0.16 {[}−0.11, +0.60{]} & +0.49 {[}+0.13, +0.80{]}* & +0.27 {[}−0.09, +0.62{]} & +0.37 {[}+0.02, +0.74{]}* & +0.36 {[}+0.00, +0.76{]}* \\
7\,d & +0.62 {[}+0.15, +1.13{]}* & +0.22 {[}−0.26, +0.91{]} & +0.62 {[}+0.20, +1.05{]}* & +0.34 {[}−0.17, +0.94{]} & +0.66 {[}+0.04, +1.42{]}* & +0.43 {[}−0.05, +0.99{]} \\
10\,d & +0.57 {[}−0.20, +1.42{]} & −0.09 {[}−0.58, +0.55{]} & +0.55 {[}−0.10, +1.19{]} & +0.65 {[}−0.16, +1.61{]} & +1.16 {[}+0.14, +2.27{]}* & +0.31 {[}−0.52, +1.24{]} \\
\end{longtable}

\Needspace{9\baselineskip}\medskip\noindent\textbf{Table 19.} HRES, alpha = 0.1: interval score (K) of each benchmark method minus that of the day-scale model, with 95\% paired intervals. Positive means the day-scale model is better; negative means the benchmark method is better; * marks intervals that exclude zero.

\nopagebreak
\begin{longtable}[]{@{}
  >{\raggedright\arraybackslash}p{(\columnwidth - 12\tabcolsep) * \real{0.0780}}
  >{\raggedright\arraybackslash}p{(\columnwidth - 12\tabcolsep) * \real{0.1537}}
  >{\raggedright\arraybackslash}p{(\columnwidth - 12\tabcolsep) * \real{0.1537}}
  >{\raggedright\arraybackslash}p{(\columnwidth - 12\tabcolsep) * \real{0.1537}}
  >{\raggedright\arraybackslash}p{(\columnwidth - 12\tabcolsep) * \real{0.1537}}
  >{\raggedright\arraybackslash}p{(\columnwidth - 12\tabcolsep) * \real{0.1537}}
  >{\raggedright\arraybackslash}p{(\columnwidth - 12\tabcolsep) * \real{0.1537}}@{}}
\toprule\noalign{}
\begin{minipage}[b]{\linewidth}\raggedright
Lead
\end{minipage} & \begin{minipage}[b]{\linewidth}\raggedright
Split (per-cell)
\end{minipage} & \begin{minipage}[b]{\linewidth}\raggedright
Mondrian season
\end{minipage} & \begin{minipage}[b]{\linewidth}\raggedright
ACI (0.005)
\end{minipage} & \begin{minipage}[b]{\linewidth}\raggedright
ACI (0.02)
\end{minipage} & \begin{minipage}[b]{\linewidth}\raggedright
Rolling 90
\end{minipage} & \begin{minipage}[b]{\linewidth}\raggedright
Rolling 365
\end{minipage} \\
\midrule\noalign{}
\endfirsthead
\toprule\noalign{}
\begin{minipage}[b]{\linewidth}\raggedright
Lead
\end{minipage} & \begin{minipage}[b]{\linewidth}\raggedright
Split (per-cell)
\end{minipage} & \begin{minipage}[b]{\linewidth}\raggedright
Mondrian season
\end{minipage} & \begin{minipage}[b]{\linewidth}\raggedright
ACI (0.005)
\end{minipage} & \begin{minipage}[b]{\linewidth}\raggedright
ACI (0.02)
\end{minipage} & \begin{minipage}[b]{\linewidth}\raggedright
Rolling 90
\end{minipage} & \begin{minipage}[b]{\linewidth}\raggedright
Rolling 365
\end{minipage} \\
\midrule\noalign{}
\endhead
\bottomrule\noalign{}
\endlastfoot
6\,h & +0.007 {[}+0.002, +0.011{]}* & −0.054 {[}−0.072, −0.037{]}* & −0.071 {[}−0.083, −0.060{]}* & −0.126 {[}−0.144, −0.112{]}* & −0.086 {[}−0.106, −0.066{]}* & −0.045 {[}−0.052, −0.039{]}* \\
12\,h & +0.004 {[}+0.001, +0.006{]}* & −0.073 {[}−0.091, −0.055{]}* & −0.063 {[}−0.075, −0.053{]}* & −0.137 {[}−0.154, −0.121{]}* & −0.081 {[}−0.102, −0.059{]}* & −0.039 {[}−0.047, −0.031{]}* \\
1\,d & +0.006 {[}+0.004, +0.007{]}* & −0.089 {[}−0.109, −0.068{]}* & −0.064 {[}−0.080, −0.049{]}* & −0.143 {[}−0.166, −0.122{]}* & −0.091 {[}−0.122, −0.061{]}* & −0.035 {[}−0.045, −0.027{]}* \\
2\,d & +0.005 {[}+0.004, +0.007{]}* & −0.099 {[}−0.120, −0.076{]}* & −0.049 {[}−0.065, −0.035{]}* & −0.112 {[}−0.135, −0.089{]}* & −0.072 {[}−0.106, −0.038{]}* & −0.026 {[}−0.034, −0.017{]}* \\
3\,d & +0.004 {[}+0.002, +0.007{]}* & −0.115 {[}−0.140, −0.084{]}* & −0.041 {[}−0.058, −0.024{]}* & −0.090 {[}−0.117, −0.063{]}* & −0.058 {[}−0.099, −0.018{]}* & −0.019 {[}−0.028, −0.010{]}* \\
5\,d & +0.004 {[}+0.000, +0.009{]}* & −0.163 {[}−0.203, −0.114{]}* & −0.027 {[}−0.051, −0.005{]}* & −0.055 {[}−0.093, −0.015{]}* & −0.035 {[}−0.099, +0.027{]} & −0.010 {[}−0.017, −0.003{]}* \\
7\,d & +0.007 {[}−0.000, +0.016{]} & −0.254 {[}−0.322, −0.170{]}* & −0.016 {[}−0.055, +0.021{]} & −0.032 {[}−0.094, +0.034{]} & −0.016 {[}−0.115, +0.089{]} & −0.002 {[}−0.010, +0.007{]} \\
10\,d & +0.010 {[}−0.004, +0.030{]} & −0.413 {[}−0.528, −0.276{]}* & +0.014 {[}−0.049, +0.073{]} & −0.001 {[}−0.115, +0.111{]} & +0.028 {[}−0.134, +0.198{]} & +0.018 {[}+0.004, +0.040{]}* \\
\end{longtable}

\end{document}